\documentclass{article}

\makeatletter
\def\input@path{{./ICLR/}}
\makeatother

\usepackage{ICLR/fancyhdr}
\usepackage{iclr2027_conference,times}

\usepackage{amsmath,amssymb,amsthm,mathtools,bm}
\usepackage{dsfont}
\usepackage{algorithm}
\usepackage{algorithmic}
\usepackage{booktabs}
\usepackage{colortbl}
\usepackage{graphicx}
\usepackage{multirow}
\usepackage{makecell}
\usepackage{wrapfig}
\usepackage{enumitem}
\usepackage{url}
\usepackage{xspace}
\usepackage{titletoc}
\usepackage{hyperref}
\usepackage[capitalize,nameinlink,noabbrev]{cleveref}

\graphicspath{{figures/}}

\definecolor{gg}{gray}{0.92}

\newcommand{\ourmodel}{RNADynNet\xspace}
\newcommand{\ourdata}{RNADynBench\xspace}

\newtheoremstyle{rnakine}{6pt}{6pt}{\normalfont}{} {\bfseries}{.}{0.5em}{}
\theoremstyle{rnakine}

\crefname{figure}{Fig.}{Figs.}
\Crefname{figure}{Fig.}{Figs.}
\crefname{table}{Tab.}{Tabs.}
\Crefname{table}{Tab.}{Tabs.}
\crefname{section}{Sec.}{Secs.}
\Crefname{section}{Sec.}{Secs.}
\crefname{equation}{Eq.}{Eqs.}
\Crefname{equation}{Eq.}{Eqs.}
\crefname{theorem}{Thm.}{Thms.}
\Crefname{theorem}{Thm.}{Thms.}
\crefname{remark}{Rem.}{Rems.}
\Crefname{remark}{Rem.}{Rems.}
\crefname{definition}{Dfn.}{Dfns.}
\Crefname{definition}{Dfn.}{Dfns.}
\crefname{algorithm}{Alg.}{Algs.}
\Crefname{algorithm}{Alg.}{Algs.}
\crefname{proposition}{Prop.}{Props.}
\Crefname{proposition}{Prop.}{Props.}

\renewcommand{\paragraph}[1]{{\vspace{0.3mm}\noindent\bfseries #1}.}

\title{RNADyn: A Benchmark for Generating and Understanding RNA Dynamics\par\vspace{0.5em}}

\author{%
\begin{tabular}{@{}c@{}}
\textbf{Yiming Huang$^{1}$ \hspace{0.8em}
Lennart Bastian$^{1}$ \hspace{0.8em}
Hanqun Cao$^{2}$ \hspace{0.8em}
Luis Vollmers$^{3}$ \hspace{0.8em}
Tolga Birdal$^{1}$}\\ [0.1em]
\normalfont $^{1}$Imperial College London \;
\normalfont $^{2}$The Chinese University of Hong Kong\\
\normalfont $^{3}$Technical University of Munich \\ [0.15em]
\normalfont \textit{Project page:} \textcolor{blue!70!black}{\url{http://circle-group.github.io/research/RNADyn}}
\end{tabular}%
}

\iclrfinalcopy

\begin{document}

\maketitle
\lhead{}

\begin{abstract}
Ribonucleic acid (RNA) functions through conformational changes that are not fully captured by static structures. However, large-scale standardized RNA dynamics data remain limited, and existing approaches typically treat trajectory generation and dynamics understanding as separate objectives.
Here, we introduce \textbf{RNADynBench}, a standardized RNA molecular dynamics (MD) benchmark with 2585 quality-controlled 100-ns all-atom trajectories and leakage-controlled splits.
Building on RNADynBench, we develop \textbf{RNADynNet}, a unified model for RNA dynamics learning that uses a shared backbone for both trajectory generation and dynamics fingerprint extraction from a single conformer.
It combines coordinate denoising, single-frame-to-trajectory alignment, and physical grounding to connect all-atom trajectory generation with dynamics representation learning.
Physical grounding improves both generated dynamics and the physical information recoverable from these fingerprints. 
Across both test sets, including the high-flexibility challenge set, the generated trajectories achieve RMSF correlations of 0.875 and 0.766, while single-conformer predictions show comparable agreement with MD-derived dynamics.
RNADynBench and RNADynNet together establish a benchmark and unified modeling framework for generating and understanding RNA dynamics.
\end{abstract}

\vspace{-2mm}
\section{Introduction}
\vspace{-2mm}
\label{sec:intro}

AI-driven structure prediction has transformed molecular biology, enabling accurate prediction of biomolecular structures directly from sequence \citep{jumper2021highly}.
Yet biological function depends not only on static structures but also on the conformational dynamics that connect multiple molecular states.
Ribonucleic acid (RNA) exemplifies this challenge due to its essential roles in regulating gene expression, recognizing molecular partners, and catalyzing biochemical reactions \citep{Ganser-NRMCB2019,gRNAde-ICLR2025}, and accurate modeling remains challenging.

On one hand, progress in RNA modeling is constrained by the limited availability and diversity of experimental structural data.
Experimentally resolved RNA structures are both less abundant and less diverse than protein structures, and only a small fraction of RNA families have been structurally characterized \citep{szikszai2024rna3db,kagaya2025nufold}.
Moreover, transformer-based architectures such as AlphaFold~\citep{jumper2021highly} rely on multiple-sequence alignments (MSAs), for which identifying and aligning sufficiently diverse RNA homologs pose challenges distinct from those in protein modeling \citep{magnus2025rnahub,kagaya2025nufold}.
While recent advances in RNA and RNA-containing complex prediction demonstrate the potential of learning-based approaches, they still leave substantial challenges regarding conformational coverage and dynamics-aware modeling \citep{kagaya2025nufold,AF3-Nature2024}.
Together, these advances and limitations underscore a need for more RNA-specific data and evaluation \citep{szikszai2024rna3db}.

\begin{figure}[t]
\vspace{-1.6em}
\centering
\setlength{\abovecaptionskip}{3pt}
\includegraphics[width=\textwidth]{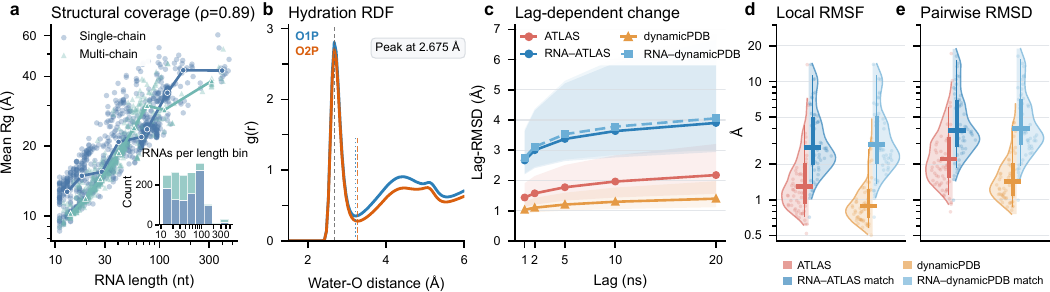}
\caption{\textbf{\ourdata visualized.}
\textbf{a}, Mean $R_\mathrm{g}$ versus RNA residue count.
Points show individual RNAs, lines connect length-binned medians, and the inset shows counts; $\rho$ is Spearman correlation between RNA length and mean $R_\mathrm{g}$.
\textbf{b}, Median phosphate--water radial distribution functions (RDFs); shading shows the interquartile range, and vertical lines mark the first peak and minima.
\textbf{c}, Weighted median lag-RMSD for size-matched RNA--protein comparisons; shading shows the weighted interquartile range (IQR).
\textbf{d},\textbf{e}, Weighted distributions of mean local RMSF (\textbf{d}) and median pairwise RMSD (\textbf{e}).
Split violins show protein left and matched RNA right, with median, IQR, and 5th--95th percentiles.
RNA uses C4$'$ atoms and protein uses C$\alpha$ atoms.
}
\vspace{-3mm}
\label{fig:data}
\end{figure}

A further challenge is dynamics.
Functionally relevant RNA states can be transient, and interactions with ions, ligands, and proteins can reshape their structural conformations \citep{Dethoff-Nature2012b,yadav2022riboswitch}.
In our size-matched comparison with the ATLAS and Dynamic PDB protein MD collections \citep{ATLAS-NAR2024,DynamicPDB-arXiv2024}, RNA trajectories show greater lag-dependent displacement and generally larger local fluctuations and pairwise RMSD over the sampled timescales (see \cref{fig:data}\textbf{c}--\textbf{e} and \cref{fig:rna-protein-stratified}).
A study of bound and unbound RNA--protein complexes likewise found that RNA often undergoes greater conformational rearrangement than proteins \citep{barik2012proteinrna}.
These observations suggest that inductive biases for modeling RNA can benefit from conditioning on molecular motion rather than on static structure alone.

All-atom molecular dynamics (MD) simulates atomic motion using forces computed from physics-based molecular potentials.
The resulting trajectories resolve conformational changes over time and allow quantification of fluctuations and correlated motions over sampled timescales \citep{sponer2018rna}.
Learning-based studies have begun to exploit such trajectories, suggesting that MD simulations can provide supervision for learning molecular dynamics.
Studies in proteins already show that incorporating dynamic and ensemble information can improve downstream prediction, motivating analogous approaches for RNA \citep{guo2025static,kalifa2026learning}.
Yet recent developments have primarily focused on proteins, 
leaving the molecular motion and folding of nucleic acids largely unexplored.
This is because \textbf{standardized RNA MD datasets remain scarce} (see \cref{tab:rna-md-resources}), and their construction requires both computational resources and expertise in molecular simulation.
Generating trajectories consumes substantial GPU resources, while preparing RNA systems, selecting force fields and solvent conditions, and establishing equilibration protocols requires domain expertise and careful physical modeling \citep{sponer2018rna}.

To address this gap, we introduce \textbf{\ourdata}, a standardized RNA MD benchmark, and \textbf{\ourmodel}, a unified model that demonstrates the capabilities enabled by such data.
\ourdata provides the training foundation for \ourmodel, supplying coordinate targets for generation, trajectory-derived targets for representation alignment, and MD-derived statistics. 
Motivated by evidence that representation guidance can improve generation \citep{REPA-ICLR2025,hgr-2026} and RNA's pronounced conformational dynamics, we introduce physical grounding (PG) to train intermediate representations to predict MD-derived covariance and inter-residue motion coupling.
Our experiments show that physical grounding improves both generated dynamics and learned dynamics representations.
Our \textbf{contributions} are fourfold:
\begin{itemize}[topsep=0em,itemsep=0mm,parsep=0pt,leftmargin=*]

\item \textbf{\ourdata.}
A standardized RNA MD benchmark with 2,585 quality-controlled 100-ns all-atom trajectories, leakage-controlled splits, and protocols for generation and dynamics prediction.

\item \textbf{\ourmodel.}
A unified framework coupling RNA trajectory generation and single-conformer dynamics learning through alignment and shared physical supervision.

\item \textbf{All-atom RNA trajectory generation benchmark.}
We establish an RNA MD generation benchmark and show that \ourmodel captures geometry, fluctuations, and conformational distributions across structural and high-flexibility tests.

\item \textbf{Transferable dynamics representations.}
Single-conformer fingerprints recover MD-derived fluctuations and motion couplings and transfer to RNA--ligand binding prediction.

\end{itemize}

\vspace{-1.5mm}
\section{\ourdata: An RNA Molecular Dynamics Benchmark}
\label{sec:rnadynbench}
\vspace{-1.5mm}
\begin{table}[t]
\vspace{-1.5em}
\centering
\caption{Comparison of RNA MD resources.}
\label{tab:rna-md-resources}
\setlength{\tabcolsep}{2pt}
\resizebox{\linewidth}{!}{%
\begin{tabular}{@{}llll@{}}
\toprule
\textbf{Resource} &
\textbf{Scale} &
\textbf{Simulation time} &
\textbf{Task} \\
\midrule
BIGNASim (2016)~\citep{BIGNASim-NAR2016} & 14 traj & 15.6~$\mu$s total & Database and analysis \\
CASP15 MD~\citep{CASP15MD-CSBJ2025} & 9 RNAs / 61 traj & 300 ns each (18.3~$\mu$s total) & Structure refinement \\
DeepRMSF~\citep{DeepRMSF-BIB2026} & 371 RNAs & 200 ns each (74.2~$\mu$s total) & RMSF prediction \\
\midrule
\rowcolor{gg}
\textbf{\ourdata}
&
\textbf{1,469 RNAs / 2,585 traj}
&
100 ns each (\textbf{258.5~$\mu$s} total)
&
\textbf{MD generation / prediction} \\
\bottomrule
\end{tabular}}
\vspace{-5mm}
\end{table}

We introduce \ourdata, a standardized benchmark for RNA trajectory generation and single-conformer dynamics prediction.
It comprises 2,585 all-atom trajectories across 1,469 RNAs, totaling 258.5~$\mu$s of simulation.
The trajectories are accompanied by molecular mappings, simulation metadata, and quality-control records, and are partitioned into leakage-controlled splits.
\ourdata defines common, architecture-agnostic evaluation protocols for both tasks.
\Cref{tab:rna-md-resources} positions its scale and intended tasks alongside existing RNA MD resources; discussed further in \cref{sec:related-work}.

\paragraph{Data Curation} 
\ourdata is constructed from experimentally resolved RNA structures retrieved from RNA-Solo, a repository of cleaned PDB-derived RNA structures \citep{RNASolo-2022}.
For each entry, we retrieve the original PDB/mmCIF records by PDB ID to recover its molecular context before curation.
We retain the RNA-Solo target chain or chains.
Additional RNA chains are retained only if their chain assignments are reliable, they contact the target within 5~\AA{} (heavy atoms) after non-RNA components are removed, and they meet the partner-size limit.
When partners are retained, the combined RNA length is limited to 512 nucleotides.
Other RNA chains, proteins, DNA, and organic ligands are removed, with component-selection decisions recorded in the metadata.
We intentionally exclude structures with unresolved backbone gaps, unsupported chemistry, or untraceable topology changes.
Missing atoms and modified nucleotides are thus only handled when residue identity and topology remain unambiguous.
Canonical atom and residue mappings link source structures, simulation topologies, and released coordinates while preserving chain identifiers, residue indices, nucleotide identities, and atom names.

\paragraph{Molecular dynamics simulation}
Each curated static structure is simulated using all-atom molecular dynamics (MD) to generate a trajectory.
MD evolves atomic coordinates by computing interatomic forces from a molecular force field and integrating Newton's equations of motion; the molecular topology remains fixed, so the simulation captures conformational motion rather than chemical reactions.
We simulate each RNA in explicit OPC water using GROMACS and the Amber RNA OL3 force field, with charge neutralization and 150 mM NaCl.
Potential-energy minimization removes unfavorable initial contacts, followed by staged equilibration under controlled temperature and pressure.
A final 5-ns NPT equilibration at 300 K and 1 bar is excluded from the released trajectories. 
Each system then undergoes a 100-ns production simulation at 300 K and 1 bar using the Nosé-Hoover thermostat \citep{Hose-Hoover-1985} and Parrinello-Rahman barostat \citep{Parrinello-1981}, with coordinates saved every 100 ps (1,001 frames).
Water and ions used in simulations are omitted from the released trajectories.
Further simulation details are given in \cref{app:rnadynbench-simulation}.

\paragraph{Quality control and benchmark splits}
We validate each simulation output by checking that the full trajectory completed successfully and by screening for coordinate jumps, backbone discontinuities, periodic-boundary artifacts, and chain separation in multi-chain systems.
Simulations showing sustained chain separation are excluded, and the separated RNA components are treated as independent systems and re-simulated.
A central challenge for RNA dynamics learning is avoiding evaluation leakage from closely related RNA structures and trajectories.
Therefore, we group related systems by RNA identity, source structure, replicate lineage, known trajectory aliases, and sequence and structural similarity, and assign each group entirely to one partition.
The \texttt{test\_struct} split evaluates structural generalization, while \texttt{test\_flex} evaluates generalization to highly flexible systems.
The release contains 2,248 training trajectories, 102 validation trajectories, 117 trajectories in \texttt{test\_struct}, and 100 in \texttt{test\_flex}, plus 18 homology-guard trajectories reserved for auditing.
Detailed quality-control and split-construction procedures are provided in \cref{app:rnadynbench-curation}.
These splits are shared by two benchmark tasks: generating 100-ns trajectories from an initial conformer and predicting MD-derived residue covariance, RMSF, and normalized motion coupling (NMC) from a single conformer.

\paragraph{Dataset characteristics}
\cref{fig:data}\textbf{a} shows broad structural coverage across RNA sizes and chain compositions, with mean $R_\mathrm{g}$ increasing strongly with RNA length ($\rho=0.89$).
Radial distribution functions (RDFs) between the non-bridging phosphate oxygens O1P/O2P and water oxygen (Ow) had the same median first-peak position of $2.675~\text{\AA}$ and median first-minimum positions of
$3.225$ and $3.275~\text{\AA}$, respectively (\cref{fig:data}\textbf{b}).
Both RDFs showed a broad secondary maximum near $4.4~\text{\AA}$.  
Median first-shell coordination numbers were 2.95 and 2.85 water molecules per non-bridging phosphate oxygen for O1P and O2P, respectively.
The first peak and minimum, together with the broad secondary maximum, are consistent with RDF patterns reported in prior nucleic-acid simulations \citep{kundu2020hydration,perepelytsya2021phosphate}, supporting the physical fidelity of \ourdata.

Beyond structural diversity, RNA in \ourdata exhibits greater conformational motion than proteins.
To control for the size dependence of RMSD and RMSF, we reweighted RNA and protein systems to a shared heavy-atom-size distribution before comparison.
Across both references, RNA showed consistently higher lag-RMSD and broader local RMSF and pairwise RMSD distributions over 1--20~ns (\cref{fig:data}\textbf{c}--\textbf{e}).
These observations highlight the need for both standardized RNA dynamics data and dynamics-aware inductive biases for RNA modeling.

\vspace{-1.5mm}
\section{\ourmodel: Unifying RNA MD Understanding and Generation}
\label{sec:method}
\vspace{-1.5mm}
\begin{figure}[t]
\vspace{-1.5em}
\centering
\setlength{\abovecaptionskip}{3pt}
\includegraphics[width=\textwidth]{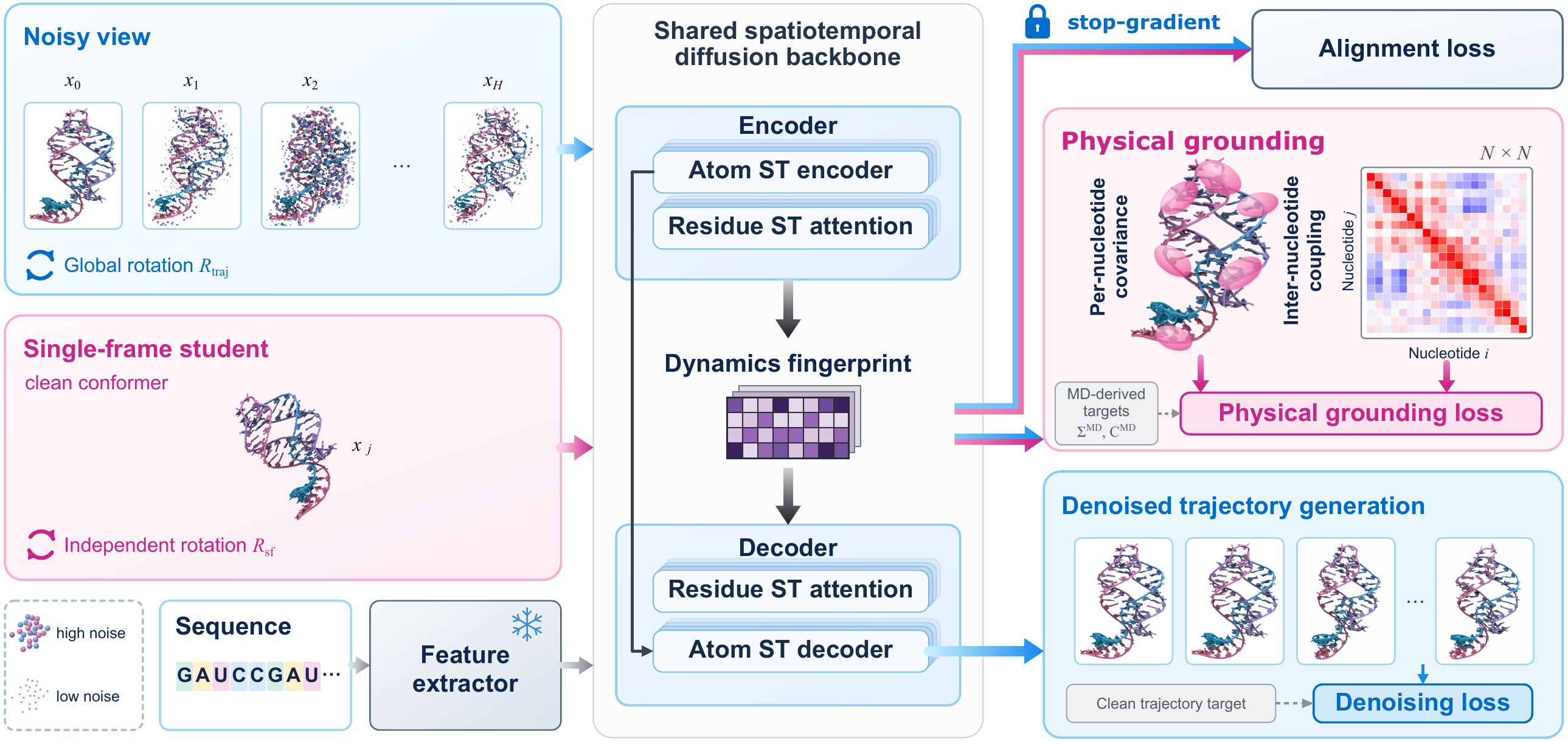}
\caption{\textbf{Overview of \ourmodel.}
\ourmodel jointly learns trajectory generation and single-conformer dynamics representations from MD trajectories.
During training, the trajectory view (blue) provides temporal context for denoising and alignment, while the single-conformer view (pink) learns a dynamics representation from one conformer.
Physical grounding supervises both views with MD-derived dynamical statistics.
At inference, a single conformer supports two modes: the denoising path generates a trajectory, whereas the encoder readout directly yields a dynamics fingerprint.
ST denotes spatiotemporal.
}
\vspace{-3mm}
\label{fig:framework}
\end{figure}

Building on \ourdata, we introduce \ourmodel, a unified model for RNA trajectory generation and single-conformer dynamics representation learning.
A shared spatiotemporal diffusion backbone supports both trajectory generation from an initial conformer and dynamics fingerprint extraction from a single conformer.
Training couples the two paths through trajectory-informed alignment and MD-derived physical supervision.
\Cref{fig:framework} summarizes our framework.

Let $\mathbf X=(\mathbf x_0,\ldots,\mathbf x_{L-1})\in\mathbb R^{L\times A\times3}$ be a trajectory window of $L$ frames, $A$ atoms, and $N$ RNA residues. 
The frames are separated by $\Delta t$ and cover a temporal horizon $H=(L-1)\Delta t$.
\ourmodel supports trajectory generation from an initial conformer $\mathbf x_0$ and deterministic encoding of a single conformer $\mathbf x$:
\begin{equation}
    \widehat{\mathbf X}_{1:L-1}
    \sim p_{\bar\theta}(\mathbf X_{1:L-1}\mid\mathbf x_0,H,\Delta t),
    \qquad
    \mathbf H_{\bar\theta}^{\mathrm{FP}}=\mathcal E_{\bar\theta}(\mathbf x).
    \label{eq:model-outputs}
\end{equation}
Here $\theta$ denotes the online parameters and $\bar\theta$ their exponential moving average (EMA). %
The distribution $p_{\bar\theta}$ is induced by the EMA denoiser and sampler, while $\mathcal E_{\bar\theta}$ denotes the shared backbone truncated at the intermediate block $\ell_{\mathrm{fp}}$ and applied to a single conformer.
Training uses $\theta$ for the single-frame pass, whereas fingerprint extraction uses $\bar\theta$.
The EMA fingerprint $\mathbf H_{\bar\theta}^{\mathrm{FP}}\in\mathbb R^{N\times d}$ is read from $\ell_{\mathrm{fp}}$; each row $\mathbf h_{\bar\theta,i}^{\mathrm{FP}}\in\mathbb R^d$ is the dynamics fingerprint of residue $i$, with $d$ the feature dimension.

\paragraph{Training objective}
\ourmodel is trained by multi-objective optimization with the losses:
\begin{equation}
\mathcal L_{\text{\ourmodel}}
=
\mathcal L_{\mathrm{gen}}
+
\lambda_{\mathrm{align}}\mathcal L_{\mathrm{align}}
+
\lambda_\Sigma\mathcal L_\Sigma
+
\lambda_C\mathcal L_C .
\label{eq:total-loss}
\end{equation}
where $\mathcal L_{\mathrm{gen}}$, $\mathcal L_{\mathrm{align}}$, $\mathcal L_\Sigma$, and $\mathcal L_C$ denote the losses for trajectory generation, single-frame-to-trajectory alignment, residue covariance, and inter-residue motion coupling, respectively.
The coefficients $\lambda_{\mathrm{align}}$, $\lambda_\Sigma$, and $\lambda_C$ weight the corresponding losses; the last two weighted terms constitute physical grounding (PG). 
Next, we provide details on the individual loss terms.

\subsection{Trajectory and Single-Frame Training}

\paragraph{Trajectory learning with heterogeneous noise}
We apply heterogeneous coordinate noise as continuous partial masking to encourage contextual learning across RNA residues and frames \citep{DiffusionForcing-NeurIPS2024}.
Heavily perturbed coordinates carry less information about the underlying conformer, whereas less perturbed positions retain more reliable structural cues.
This information asymmetry encourages the denoiser to use cleaner spatial and temporal context to reconstruct noisier regions, thereby exploiting dependencies across residues and frames.

Noise scales are assigned at frame--residue positions and shared by the atoms of each residue, forming the tensor $\boldsymbol{\sigma}$, such that $\widetilde{\mathbf X} =
\mathbf R_{\mathrm{traj}}
\left(\mathbf X + \boldsymbol{\sigma}\odot\boldsymbol\epsilon \right)$ where $\boldsymbol\epsilon\sim\mathcal{N}(\mathbf0,\mathbf I)$.
Each window is aligned and centered before corruption, and one global rotation $\mathbf R_{\mathrm{traj}}\in\mathrm{SO}(3)$ acts on coordinates and noise across all frames.
Training windows vary in start, duration, and sampling interval; observed frames use a near-clean noise scale for conditioning.
The denoiser reconstructs the clean trajectory in the same rotated coordinate frame with the \emph{coordinate-denoising loss}:
\begin{equation}
\mathcal L_{\mathrm{denoise}}
=
\mathbb E
    \left\|
        \mathcal D_\theta
        (\widetilde{\mathbf X},\boldsymbol{\sigma},H,\Delta t)
        -
        \mathbf R_{\mathrm{traj}}\mathbf X
    \right\|_w^2.
\label{eq:denoise}
\end{equation}
The weighted coordinate MSE $\|\cdot\|_w^2$ covers valid atoms in all sampled frames, including observed frames, and incorporates the EDM and generative-group weights.
The generation objective augments this loss with auxiliary supervision for local structure, atomic and pairwise-distance fluctuations, and short-lag temporal correlations.
Its terms and weights are specified in \cref{app:training-losses}.
These auxiliary losses act on generated coordinates, whereas PG supervises intermediate representations.

\paragraph{Single-frame-to-trajectory alignment}
The goal is to infer trajectory-level information from a single experimentally available conformer, rather than merely encode its static geometry.
MD trajectories provide temporal context during training, but this context is unavailable when only one conformer is observed.
We therefore train the single-conformer encoder to predict residue features from the trajectory pass, transferring information about RNA motion into its latent states.

For each forward window, we sample an anchor frame $J\sim\mathrm{Uniform}\{0,\cdots, L-1\}$ to pair the trajectory and single-frame views at the same conformer $\mathbf x_J$.
Both training states below correspond to this anchor; we omit $J$ from their notation.
From the online trajectory pass, we read the residue state $\mathbf h_{\theta,i}^{\mathrm{traj}}$ at intermediate block $\ell_{\mathrm{fp}}$ for frame $J$.
Motivated by layer-dependent representation transferability \citep{InformationFlow-ICML2025}, we use $\ell_{\mathrm{fp}}=17$ in the reported model and evaluate an alternative Block~24 readout and supervision depth in \cref{app:ablations}.
This state incorporates information from the full trajectory window through temporal attention.
After normalization and stop-gradient, it gives the alignment target $\mathbf z_i^{\mathrm{traj}}$.
A separate pass through the shared online encoder receives the clean conformer $\mathbf R_{\mathrm{sf}}\mathbf x_J$, producing $\mathbf H_\theta^{\mathrm{FP}}
=\mathcal E_\theta(\mathbf R_{\mathrm{sf}}\mathbf x_J)$, whose $i$th row is $\mathbf h_{\theta,i}^{\mathrm{FP}}$.
The rotations $\mathbf R_{\mathrm{traj}}$ and $\mathbf R_{\mathrm{sf}}$ are sampled independently and uniformly from $\mathrm{SO}(3)$.
Although the backbone is not explicitly SE(3)-equivariant, centering and rotation augmentation encourage equivariant trajectory generation and orientation-invariant dynamics fingerprints.

We map the single-frame states to the trajectory targets using a predictor $q_\psi$  conditioned on $H$ and $\Delta t$:
\begin{equation}
    \mathcal L_{\mathrm{align}}
    =
    1-\mathbb{E}_{J,i}
        \cos \left(
            q_\psi(\mathbf h_{\theta,i}^{\mathrm{FP}},H,\Delta t),
            \mathbf z_i^{\mathrm{traj}}
        \right).
    \label{eq:alignment-loss}
\end{equation}
Here $\cos$ denotes cosine similarity, and the expectation averages over sampled anchors and valid residues.
Alignment gradients update the single-frame encoder and predictor, while the trajectory target is detached.
The target reuses the existing online trajectory pass, requiring no separate target network or additional trajectory pass.
Fingerprint extraction with $\mathcal E_{\bar\theta}$ requires neither $H$ nor $\Delta t$.

\paragraph{Backbone implementation}
The backbone uses an AF3-style spatial denoiser \citep{AF3-Nature2024} implemented and initialized with Protenix \citep{protenix-2025}, with the upstream feature extractor (Input Embedder and Pairformer) frozen.
Static features from the frozen Input Embedder and Pairformer are provided to the backbone but omitted from network arguments for notational simplicity.
Each spatiotemporal block applies spatial attention followed by temporal self-attention with a Langevin-inspired temporal decay bias \citep{BioKinema-bioRxiv2026}.

\vspace{-1mm}
\subsection{Physical grounding}
\label{sec:physical-grounding}
\vspace{-1mm}
Alignment transfers trajectory-contextualized information, but does not explicitly tie the learned representations to physical observables.
Motivated by evidence that representation guidance can improve generation \citep{REPA-ICLR2025} and RNA's pronounced conformational dynamics, we use PG to explicitly shape intermediate representations with MD-derived residue covariance and inter-residue motion coupling.
PG losses update the shared backbone through both trajectory-anchor and single-conformer representations.

\paragraph{MD-derived physical targets}
We use residue covariance and inter-residue motion coupling as complementary descriptors of RNA dynamics \citep{DynaProt-ICLR2026}.
Targets are computed from a fixed $101$-frame, $100\,\mathrm{ns}$ reference trajectory, independently of the sampled training-window duration.
Each residue is represented by its C1$'$ position, and all reference frames are rigidly aligned to the initial conformer.
Writing $\delta\mathbf r_i=\mathbf r_i-\langle\mathbf r_i\rangle$, we define
\begin{equation}
\boldsymbol\Sigma_i
=\langle\delta\mathbf r_i\delta\mathbf r_i^\top\rangle,\quad
\mathrm{RMSF}_i
=\sqrt{\operatorname{tr}(\boldsymbol\Sigma_i)},\qquad
C_{ij}
=\langle\delta\mathbf r_i^\top\delta\mathbf r_j\rangle,\quad
\rho_{ij}
=C_{ij}/\sqrt{C_{ii}C_{jj}} .
\label{eq:pg-targets}
\end{equation}
Here $\langle\cdot\rangle$ denotes the average over reference conformations, $\boldsymbol\Sigma_i$ captures residue-level fluctuation magnitude and direction, while $C_{ij}$ and $\rho_{ij}$ describe pairwise motion coupling and its normalized form (NMC), respectively.
The labels satisfy $C_{ii}=\operatorname{tr}(\boldsymbol\Sigma_i)$. These MD statistics serve as supervision, not encoder inputs.

\paragraph{Physical grounding losses}
For residue covariance, a shared MLP head $f_\Sigma$ predicts the six independent entries of a symmetric $3\times3$ log-covariance matrix.
For residue $i$, we construct an orthonormal frame $\mathbf R_i^{(J)}$ from the clean anchor conformer $\mathbf x_J$ and express the $\boldsymbol\Sigma_i$ in this frame:
$\boldsymbol\Sigma_{i,\mathrm{loc}}^{(J)}
=
(\mathbf R_i^{(J)})^\top
\boldsymbol\Sigma_i
\mathbf R_i^{(J)}$.
For each view $v\in\{\mathrm{FP},\mathrm{traj}\}$, let $\mathbf h_{\theta,i}^{v}$ denote the online single-conformer or corresponding trajectory-anchor state.
Both views use the same head and clean-anchor target:
\begin{equation}
\mathcal L_\Sigma
=
\frac{1}{2|\mathcal V|}
\sum_{v\in\{\mathrm{FP},\mathrm{traj}\}}
\sum_{i\in\mathcal V}
\left\|
f_\Sigma(\mathbf h_{\theta,i}^{v})
-
\log\left(
\boldsymbol\Sigma_{i,\mathrm{loc}}^{(J)}
\right)
\right\|_{\mathrm{F},\mathrm{mean}}^2 .
\label{eq:pg-covariance}
\end{equation}
Here $\mathcal V$ contains the valid residues and $\|\cdot\|_{\mathrm{F},\mathrm{mean}}^2$ is the mean squared matrix-entry error.
Matrix logarithms act on covariance values expressed numerically in \AA$^2$.
At inference, the matrix exponential yields a positive-definite covariance, recovered in global coordinates as
$
\widehat{\boldsymbol\Sigma}_i
=
\mathbf R_i
\exp
\bigl(
f_\Sigma(\mathbf h_{\bar\theta,i}^{\mathrm{FP}})
\bigr)
\mathbf R_i^\top
$.
Local-frame construction and numerical details are given in \cref{app:physical-grounding}.

For inter-residue motion coupling, the head $f_C$ predicts a symmetric positive-definite motion-coupling matrix $\widehat{\mathbf C}$ from intermediate residue states.
Following \cite{BackFlip2-arXiv2026}, we supervise its matrix pattern, magnitude, and normalized off-diagonal coupling with
\begin{equation}
\mathcal L_C
=
1-
\frac{\langle\widehat{\mathbf C},\mathbf C\rangle_F}
{\|\widehat{\mathbf C}\|_F\|\mathbf C\|_F}
+
\frac{\alpha}{|\mathcal{V}|^2}
\sum_{i,j}
|\widehat C_{ij}-C_{ij}| 
+
\frac{\beta}{|\mathcal{V}|(|\mathcal{V}|-1)}
\sum_{i\ne j}
(\widehat\rho_{ij}-\rho_{ij})^2 .
\label{eq:coupling-loss}
\end{equation}
Here $\alpha$ and $\beta$ weight the magnitude and normalized off-diagonal coupling terms, respectively.
The construction of $f_C$ and numerical details are given in \cref{app:physical-grounding}.

At inference, trajectory generation uses the EMA denoiser with a hierarchical forecasting-and-interpolation sampler, with alignment and PG heads omitted. Single-conformer inference uses the encoder alone for fingerprint extraction, with PG heads used only for direct descriptor prediction.

\begin{table}[t]
    \vspace{-1.5em}
    \centering
    \caption{Trajectory generation on \ourdata. Bold marks the best.}
    \label{tab:generation}
    \scriptsize
    \setlength{\tabcolsep}{2pt}
    \resizebox{\linewidth}{!}{%
    \begin{tabular}{@{}lrrrrrrr@{\hspace{4pt}}rrrrrrr@{}}
        \toprule
        \multirow{2}{*}[-2.6ex]{Method} & \multicolumn{7}{c}{\texttt{test\_struct}} & \multicolumn{7}{c}{\texttt{test\_flex}} \\
        \cmidrule(lr){2-8}\cmidrule(l){9-15}
         & \makecell{Bond\\MAE$\downarrow$} & \makecell{O3P\\max$\downarrow$} & \makecell{Chiral\\flip$\downarrow$} & \makecell{Clash\\/1k$\downarrow$} & \makecell{RMSF\\$r\uparrow$} & \makecell{PCA\\$W_2\downarrow$} & \makecell{GPU\\time} & \makecell{Bond\\MAE$\downarrow$} & \makecell{O3P\\max$\downarrow$} & \makecell{Chiral\\flip$\downarrow$} & \makecell{Clash\\/1k$\downarrow$} & \makecell{RMSF\\$r\uparrow$} & \makecell{PCA\\$W_2\downarrow$} & \makecell{GPU\\time} \\
        \midrule
        Reference & 0.03308 & 0.1938 & 0 & 0 & 0.9819 & 0.7295 & $8.1 \mathrm{h}$ & 0.03307 & 0.1966 & 0 & 0 & 0.9633 & 3.1137 & $13.9\mathrm{h}$ \\
        ConfRover & 0.19315 & 120.4747 & 0.03103 & 581.4520 & 0.0666 & 9.3105 & 340.4s & 0.16506 & 133.0420 & 0 & 419.7848 & 0.2288 & 11.1922 & 501.5s \\
        MDGen & 0.08024 & 15.1488 & 0.03103 & 95.9398 & 0.3737 & 2.4200 & 7.2s & 0.10786 & 18.9945 & $\mathbf{<10^{-5}}$ & 88.9058 & 0.2994 & 8.9282 & 11.0s \\
        BioKinema & 0.02612 & 4.4682 & 0.03836 & 2.0746 & \textbf{0.8770} & 1.9276 & 40.2s & 0.02950 & 8.3586 & 0.02012 & 5.8241 & 0.7536 & 5.9406 & 47.8s \\
        \midrule
        w/o PG & \textbf{0.02418} & 2.4455 & 0.03097 & 1.1884 & 0.8594 & 1.9085 & / & \textbf{0.02555} & 9.1171 & 0.00160 & \textbf{2.8416} & 0.7440 & 6.1744 & / \\
        \ourmodel{} & 0.02424 & \textbf{1.8858} & \textbf{0.03085} & \textbf{0.8225} & 0.8746 & \textbf{1.8739} & 24.5s & 0.02567 & \textbf{4.2400} & 0.00093 & 3.0404 & \textbf{0.7658} & \textbf{5.9169} & 32.9s \\
        \bottomrule
    \end{tabular}%
    \vspace{-1cm}
    }
\end{table}

\vspace{-2mm}
\section{Experiments}
\label{sec:experiments}
\vspace{-2mm}

We evaluate \ourmodel on three tasks: trajectory generation, single-conformer dynamics prediction, and RNA--ligand binding transfer.
Generation and dynamics prediction use the standard \texttt{test\_struct} and \texttt{test\_flex} splits from \ourdata.
Experimental protocols and supplementary results are provided in \cref{app:experimental-setup,app:additional-results}.

\vspace{-1.5mm}
\subsection{RNA trajectory generation under dynamics-aware evaluation}
\label{sec:exp-generation}
\vspace{-1.5mm}
Existing molecular dynamics generation methods have primarily been developed for proteins rather than RNA.
We therefore adapt ConfRover~\citep{ConfRover-NeurIPS2025}, MDGen~\citep{MDGen-NeurIPS2024}, and BioKinema~\citep{BioKinema-bioRxiv2026} to RNA and retrain them on \ourdata under the same 12h training limit.
The adaptations are described in \cref{app:baseline-details}.
Each method receives only the initial conformer and generates 101-frame trajectories spanning 100 $\mathrm{ns}$.
For each test initial conformer, we generate four trajectories, except ConfRover ($K=1$) due to its slow sampling.
The Reference row reports MD-derived metric values and MD simulation time.
\Cref{tab:generation} depicts local geometric quality, agreement with MD dynamics, and sampling efficiency.
We report bond and O3$'$--P errors, sugar-chirality flips, steric clashes, and MD-reference PCA-$W_2$ distance.
We also report Pearson correlation across residues between generated-sample-mean and MD C1$'$ RMSF profiles.
Bond, O3$'$--P, and chirality metrics use the initial conformer as reference.
Distances are in \AA{}, flips in percent, clashes per 1,000 heavy atoms, and runtimes are measured per trajectory on H200 GPUs.
Metric details are given in \cref{app:evaluation-metrics}. 

\ourmodel improves all four reported geometric metrics over BioKinema on both test sets, while achieving comparable or better RMSF correlation, lower PCA-$W_2$, and faster sampling.
In particular, the maximum O3$'$--P deviation and clash rate are reduced by roughly half on both test sets.
MDGen samples faster but shows substantially larger backbone deviations, more clashes, and weaker dynamic fidelity.
ConfRover performs worse on most metrics and is much slower to sample; its lower training throughput under the common 12h budget may also have limited convergence.

\begin{wrapfigure}[29]{r}{0.5\textwidth}
\centering
\includegraphics[width=\linewidth]{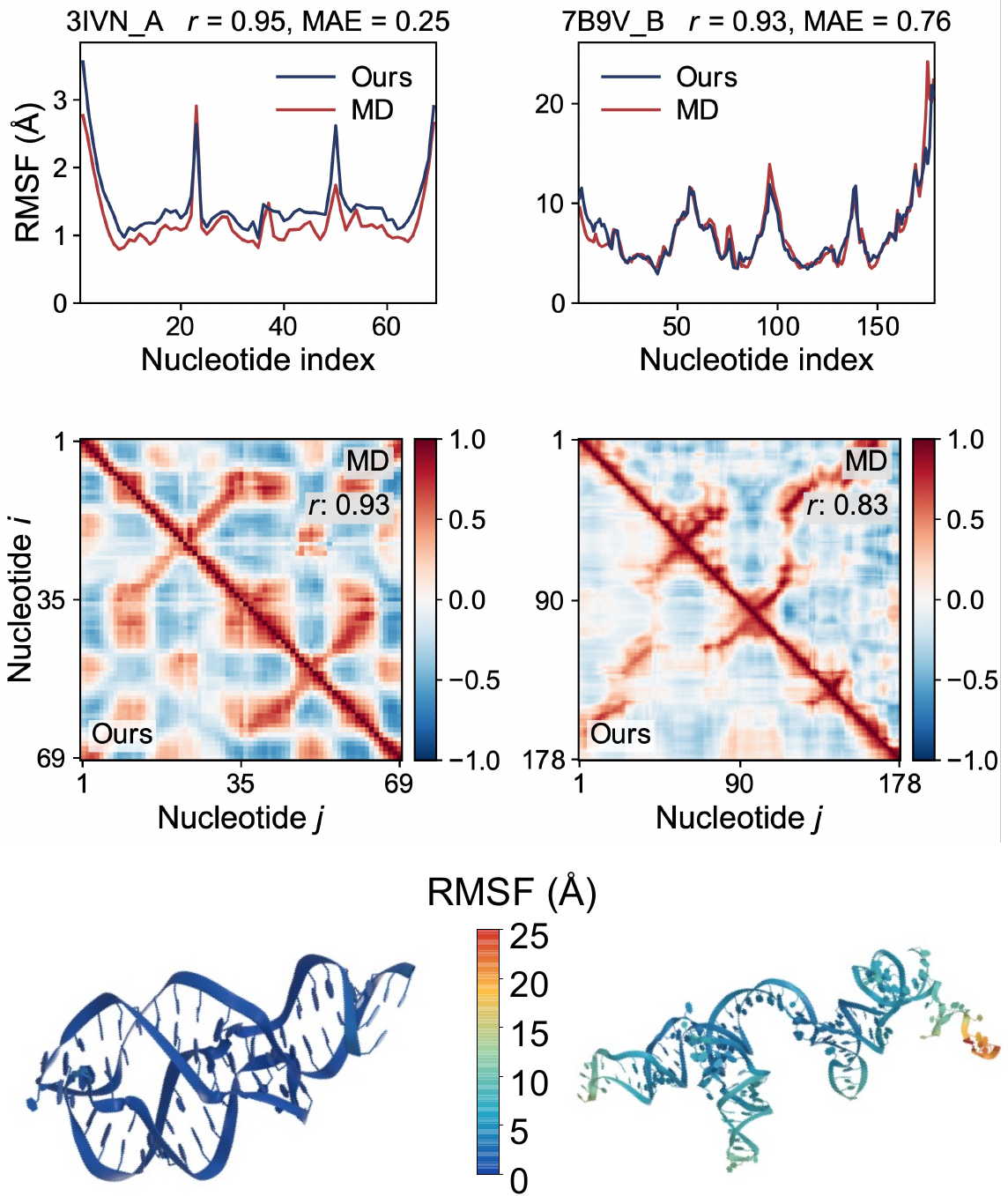}
\caption{Dynamics inference from single conformers.
Predicted per-nucleotide RMSF and pairwise nucleotide motion couplings
from \ourmodel are compared with MD references for two RNAs.
Motion-coupling matrices show \ourmodel below the diagonal and MD above; structures are colored by RMSF.}
\label{fig:dynamics-prediction}
\end{wrapfigure}
Physical grounding also improves generation.
Relative to w/o PG, the full model improves RMSF correlation, O3$'$--P continuity, and PCA-$W_2$ on both panels, with similar bond MAE. 
Clash rates decrease on \texttt{test\_struct} and change modestly on \texttt{test\_flex}. 
Overall, MD-derived physical supervision improves local backbone geometry and conformational dynamics.

\vspace{-1mm}
\subsection{Learning RNA dynamics\\from single conformers}
\label{sec:dynamics-prediction}
\vspace{-1mm}
A central goal of \ourdata is to enable learning of molecular dynamics properties from experimentally available static conformers.
We therefore evaluate whether \ourmodel can recover MD-derived dynamics without trajectory generation at inference.

Models receive a single RNA conformer and predict MD-derived residue covariance, RMSF, and NMC.
Native heads measure direct prediction accuracy, whereas independently trained probes on frozen encoders measure the dynamics information contained in learned representations.
\Cref{tab:dynamics} compares \ourmodel and its ablations with RNA-FM~\citep{RNAFM-2022}, Geometry-GNN, RNA-FM+2D GCN, BackFlip~\citep{BackFlip2-arXiv2026}, and DynaProt~\citep{DynaProt-ICLR2026}.
RMWD and symmetric KL quantify overall covariance mismatch, while ellipsoid IoU measures agreement in fluctuation anisotropy and orientation.
RMSF and NMC capture per-residue fluctuation magnitude and pairwise motion coupling, respectively, and are evaluated by Pearson $r$ and MAE.
Details are given in \cref{app:evaluation-metrics}.

\begin{table}[t]
    \vspace{-5.0mm}
    \centering
    \caption{Single-conformer prediction of MD-derived RNA dynamics. The MD reference compares two non-overlapping 50-ns MD blocks; model predictions are evaluated against full 100-ns MD statistics. $\dagger$ denotes probes on frozen representations. Bold denotes the best result.}
    \label{tab:dynamics}
    \scriptsize
    \setlength{\tabcolsep}{2pt}
    \resizebox{\linewidth}{!}{%
    \begin{tabular}{@{}lrrrrrrr@{\hspace{1pt}}rrrrrrr@{}}
        \toprule
        & \multicolumn{7}{c}{\texttt{test\_struct}} & \multicolumn{7}{c}{\texttt{test\_flex}} \\
        \cmidrule(lr){2-8}\cmidrule(l){9-15}
        & \multicolumn{2}{c}{RMSF} & \multicolumn{3}{c}{Covariance} & \multicolumn{2}{c}{NMC}
        & \multicolumn{2}{c}{RMSF} & \multicolumn{3}{c}{Covariance} & \multicolumn{2}{c}{NMC} \\
        \cmidrule(lr){2-3}\cmidrule(lr){4-6}\cmidrule(lr){7-8}\cmidrule(lr){9-10}\cmidrule(lr){11-13}\cmidrule(l){14-15}
        Method & $r\uparrow$ & MAE$\downarrow$ & RMWD$\downarrow$ & Sym. KL$\downarrow$ & IoU$\uparrow$ & $r\uparrow$ & MAE$\downarrow$ & $r\uparrow$ & MAE$\downarrow$ & RMWD$\downarrow$ & Sym. KL$\downarrow$ & IoU$\uparrow$ & $r\uparrow$ & MAE$\downarrow$ \\
        \midrule
        MD reference & 0.8755 & 0.3770 & 0.7960 & 0.5525 & 0.5497 & 0.8532 & 0.1462 & 0.7584 & 1.6415 & 3.5505 & 1.3471 & 0.4185 & 0.7520 & 0.2145 \\
        \midrule
        RNA-FM$^{\dagger}$ & 0.2939 & 1.6188 & 2.1173 & 4.9868 & 0.1731 & 0.3690 & 0.2644 & 0.2085 & 3.8148 & 5.0390 & 4.2367 & 0.2318 & 0.3635 & 0.2952 \\
        Geometry-GNN & 0.2801 & 1.1241 & 1.5796 & 2.8662 & 0.2561 & 0.4821 & 0.2591 & 0.3385 & 3.4276 & 4.6870 & 3.1959 & 0.3054 & 0.4442 & 0.2910 \\
        RNA-FM+2D GCN & 0.1257 & 2.0117 & 2.6303 & 8.5914 & 0.1831 & 0.4797 & 0.2569 & 0.1536 & 3.5511 & 4.8377 & 4.9398 & 0.2625 & 0.5333 & 0.2609 \\
        BackFlip & 0.8098 & 0.4608 & 0.9449 & 0.7599 & 0.4919 & 0.7948 & 0.1707 & 0.7311 & 1.7974 & 3.7118 & 4.4011 & 0.4058 & 0.7399 & 0.2084 \\
        DynaProt & 0.7726 & 0.5266 & 0.9786 & 0.8288 & 0.4502 & 0.2454 & 0.2775 & 0.5518 & 2.5861 & 3.9757 & 1.9398 & 0.3692 & 0.1889 & 0.3132 \\
        \midrule
        w/o align. \& PG$^{\dagger}$ & 0.7473 & 0.6056 & 1.0756 & 0.9827 & 0.3966 & 0.7924 & 0.1717 & 0.6232 & 2.1515 & 3.8137 & 1.3343 & 0.3811 & 0.7506 & 0.2088 \\
        w/o PG$^{\dagger}$ & 0.7797 & 0.5160 & 0.9839 & 0.7542 & 0.4540 & 0.7834 & 0.1744 & 0.6920 & 1.8983 & 3.5810 & 1.1055 & 0.4069 & 0.7443 & 0.2101 \\
        \ourmodel{}$^{\dagger}$ & 0.8626 & 0.4076 & 0.8741 & 0.5823 & \textbf{0.5276} & 0.7958 & 0.1694 & 0.7782 & 1.7560 & 3.3359 & 0.8775 & 0.4522 & 0.7577 & 0.2035 \\
        \ourmodel{} & \textbf{0.8665} & \textbf{0.3983} & \textbf{0.8651} & \textbf{0.5743} & 0.5260 & \textbf{0.8115} & \textbf{0.1629} & \textbf{0.7827} & \textbf{1.5800} & \textbf{3.2700} & \textbf{0.8533} & \textbf{0.4528} & \textbf{0.7614} & \textbf{0.2027} \\
        \bottomrule
    \end{tabular}%
    \vspace{-1cm}
    }
\end{table}

\ourmodel's native heads outperform BackFlip and DynaProt on all
seven reported metrics in both test panels.
RMSF correlations reach $0.8665$ and $0.7827$, while NMC correlations reach $0.8115$ and $0.7614$ on \texttt{test\_struct} and \texttt{test\_flex}, respectively.
\Cref{fig:dynamics-prediction} illustrates predicted RMSF and motion couplings for two RNAs.
Frozen-probe ablations show that alignment improves all RMSF and covariance metrics on both test sets, although NMC is slightly worse.
Adding physical grounding improves all seven metrics on both sets. 
These results suggest that alignment transfers trajectory-derived information to the single-conformer representation, while physical grounding strengthens its recovery of MD-derived dynamics.

\begin{wrapfigure}[17]{r}{0.46\textwidth}
\centering
\vspace{-3.5mm}
\includegraphics[width=\linewidth]{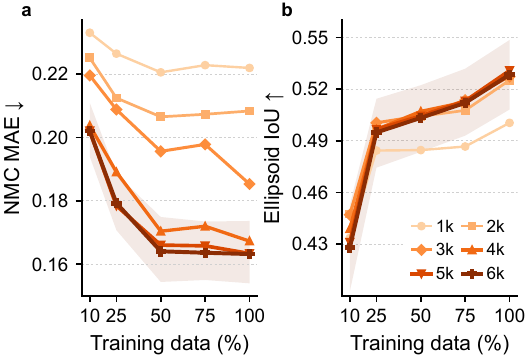}
\vspace{-5mm}
\caption{\textbf{Predictive data scaling} on \texttt{test\_struct}.
NMC MAE (\textbf{a}) and covariance-ellipsoid IoU (\textbf{b}) across 10--100\% of the training data and 1k--6k training steps.}
\label{fig:data-scaling}
\end{wrapfigure}

\noindent\textbf{Scaling RNA dynamics learning with MD supervision.}
To examine how trajectory data contribute to dynamics learning, we train \ourmodel{} on five fractions of the training data spanning 10--100\% and evaluate checkpoints from 1k to 6k steps (\cref{fig:data-scaling}).
At later checkpoints, larger subsets generally reduce NMC MAE and increase ellipsoid IoU, suggesting that additional MD supervision improves the recovery of RNA dynamics properties.

\vspace{-3mm}
\subsection{RNA--ligand binding}
\label{sec:exp-rna-ligand-binding}
\vspace{-3mm}
If learned dynamics fingerprints capture biologically relevant conformational information, they should improve tasks where RNA
recognition depends on structural adaptation.
We therefore assess downstream transfer to RNA--ligand binding using frozen \ourmodel fingerprints in the GerNA-Bind framework
~\citep{GerNABind-NMI2025}.

\Cref{tab:binding-auroc} shows that the binding model using frozen \ourmodel fingerprints improves AUROC in five of eight settings.
Notably, performance improves on both RNA-homology splits and both combined homology-and-fingerprint splits, with the latter increasing from $0.7279$ to $0.7673$ on Biosensor and from $0.6176$ to $0.6619$ on Robin.
Overall, the gains are concentrated on settings involving RNA distribution shift.

\begin{table}[t]
    \vspace{-5mm}
    \centering
    \caption{RNA--ligand binding AUROC (\textbf{best}, \underline{second best}).}
    \label{tab:binding-auroc}
    \setlength{\tabcolsep}{3pt}
    \resizebox{\textwidth}{!}{%
    \begin{tabular}{@{}lcccccccc@{}}
        \toprule
        & \multicolumn{4}{c}{AUROC-Biosensor} & \multicolumn{4}{c}{AUROC-Robin} \\
        \cmidrule(lr){2-5}\cmidrule(lr){6-9}
        Method & Random & \makecell{RNA\\homology} & \makecell{Ligand\\fingerprint} & \makecell{Homol. \&\\fingerpr.} & Random & \makecell{RNA\\homology} & \makecell{Ligand\\fingerprint} & \makecell{Homol. \&\\fingerpr.} \\
        \midrule
        RSAPred~\citep{RSAPred-BIB2024} & 0.8764 & 0.7550 & 0.6707 & 0.6019 & 0.6327 & 0.5392 & 0.6320 & 0.4938 \\
        DeepDTIs~\citep{DeepDTIs-JPR2017} & 0.9300 & 0.8399 & 0.6813 & 0.6118 & 0.6302 & 0.5503 & 0.6290 & 0.4987 \\
        DeepConv-DTI~\citep{DeepConvDTI-PLoSCB2019} & 0.9249 & 0.8427 & 0.6894 & 0.6213 & 0.6301 & 0.5625 & 0.6390 & 0.5104 \\
        GraphDTA~\citep{GraphDTA-Bioinformatics2021} & 0.8992 & 0.8284 & 0.7014 & 0.6370 & 0.6590 & 0.5528 & 0.6481 & 0.5510 \\
        GerNA-Bind~\citep{GerNABind-NMI2025} & \textbf{0.9755} & \underline{0.9014} & \textbf{0.7723} & \underline{0.7279} & \underline{0.7094} & \underline{0.6188} & \textbf{0.6814} & \underline{0.6176} \\
        \midrule
        \ourmodel{} & \underline{0.9582} & \textbf{0.9066} & \underline{0.7041} & \textbf{0.7673} & \textbf{0.7119} & \textbf{0.6899} & \underline{0.6760} & \textbf{0.6619} \\
        \bottomrule
    \end{tabular}%
    \vspace{-7mm}
    }
\end{table}

\vspace{-2mm}
\section{Related Work}
\label{sec:related-work}
\vspace{-2mm}
\paragraph{Biomolecular MD Data}
Protein dynamics learning benefits from larger MD resources, including BioEmu for protein equilibrium ensembles, mdCATH for protein trajectories, and MISATO for protein--ligand MD data \citep{BioEmu-Science2025,mdCATH-SciData2024,MISATO-NCS2024}.
ATLAS and Dynamic PDB provide protein MD resources with flexibility descriptors and additional physical properties, respectively \citep{ATLAS-NAR2024,DynamicPDB-arXiv2024}.
RNA trajectory resources remain smaller and more specialized (\cref{tab:rna-md-resources}).
BIGNASim is a database and analysis portal for nucleic-acid MD trajectories, initially including 14 RNA simulations \citep{BIGNASim-NAR2016}.
CASP15 MD evaluates refinement of predicted RNA structures \citep{CASP15MD-CSBJ2025}, while the simulations assembled for DeepRMSF support flexibility prediction \citep{DeepRMSF-BIB2026}.
\ourdata instead contributes a larger all-atom RNA trajectory collection with RNA-group partitions and common protocols for generating and predicting MD-derived descriptors.

\paragraph{Ensemble and Trajectory Generation}
Protein ensemble generators such as AlphaFlow~\citep{AlphaFlow-ICML2024} and BioEmu~\citep{BioEmu-Science2025}, and RNA generators such as DynaRNA~\citep{DynaRNA-CB2025} and RNAbpFlow~\citep{RNAbpFlow-NM2026}, sample conformations without temporal order.
For time-resolved dynamics, ConfRover models protein dynamics autoregressively~\citep{ConfRover-NeurIPS2025}, whereas MDGen jointly models full trajectories in an SE(3)-invariant representation of residue poses and torsions~\citep{MDGen-NeurIPS2024}.
BioMD and BioKinema instead operate directly on all-atom Cartesian trajectories using spatiotemporal models and hierarchical forecasting and interpolation, with BioKinema additionally introducing a Langevin-inspired temporal decay bias~\citep{BioMD-ICLR2026,BioKinema-bioRxiv2026}.
Nevertheless, none of these trajectory generators was originally developed for RNA MD generation.

\paragraph{Dynamics Prediction from Structure}
RNA elastic networks and DeepRMSF estimate flexibility from a single structure \citep{Pinamonti-NAR2015,DeepRMSF-BIB2026}.
In proteins, DynaProt predicts directional residue fluctuations and pairwise motion couplings~\citep{DynaProt-ICLR2026}, while Ensembits distills information from MD ensembles into representations accessible from a single frame~\citep{Ensembits-arXiv2026}.
\ourmodel extends RNA structure-based dynamics prediction beyond scalar flexibility by learning MD-derived residue covariance and inter-residue motion coupling from a single conformer, within the same backbone used for trajectory generation.

\paragraph{Guided Generative and Representation Learning}
Recent work has explored incorporating auxiliary signals into generative and representation learning.
REPA aligns generative features with pretrained representations~\citep{REPA-ICLR2025}, HoG-Diff guides graph diffusion with higher-order topology~\citep{hogdiff-ICLR2026}, and PSG-JEPA grounds predictive latents with physical state and multi-horizon transition targets~\citep{PSGJEPA-arXiv2026}.
We instead use MD trajectories to jointly supervise trajectory generation and single-conformer dynamics representations through trajectory-informed alignment and MD-derived physical targets.

\vspace{-1.5mm}
\section{Conclusion}
\label{sec:conclusion}
\vspace{-1.5mm}
\ourdata establishes the data foundation for jointly learning RNA trajectory generation and single-conformer dynamics prediction.
Building on its trajectories and MD-derived statistics, \ourmodel integrates these capabilities within a shared backbone.
Physical grounding improves generated dynamics and the physical information recoverable from single-conformer fingerprints, while an initial RNA--ligand binding evaluation illustrates their downstream transfer.
\ourdata provides a foundation for moving RNA modeling beyond static structure toward models that learn conformational dynamics as a bridge between molecular structure, recognition, and function.

\paragraph{Limitations and future work}
Our simulations span 100~ns and may not capture slower conformational dynamics.
The study focuses primarily on isolated RNA and does not explicitly model complete RNA--protein or RNA--small-molecule complexes.
In the future, \ourdata could support dynamics-aware RNA design, perturbation-response prediction, and integration with experimental observables.
It could also enable studies of how mutations, molecular partners, and environmental conditions reshape RNA dynamics. %

\clearpage
\subsection*{AI use statement}
In this work, we used generative AI tools for language editing and for limited code editing, debugging, and visualization. All AI-assisted text and code were reviewed by the authors. The authors take full responsibility for the final content and artifacts of this work.

\subsection*{Ethics and Reproducibility statement}
RNA dynamics modeling has many promising applications benefiting humanity, including understanding RNA function, molecular recognition, and RNA-targeting therapeutics.
More capable models could also support the engineering of functional RNAs or RNA--molecule interactions, raising potential dual-use concerns.
RNADynNet does not perform de novo RNA design or establish biological activity.
Therefore, RNA dynamics models generally need to be applied with caution when used in molecular design or therapeutic pipelines.

\cref{app:rnadynbench-details}-\cref{app:experimental-setup} describe data curation, MD simulation and quality control, model architecture and training, baseline adaptation, evaluation metrics, and downstream experiments.
\cref{app:additional-results} reports additional scaling, ablation, and statistical results.

\subsection*{Acknowledgments}
TB acknowledges support from the UKRI Engineering and Physical Sciences Research Council (EPSRC) through the Future Leaders Fellowship [grant number MR/Y018818/1].
YH, LB, and TB are grateful for support from the UK AI Research Resource (AIRR) through grant 0251-4584-0945-1.
YH acknowledges support from the Lee Family Scholarship at Imperial College London.
LB was supported by the UK Royal Society through grant NIF/R1/254128.

\clearpage
\appendix
\startcontents[appendix]
\begin{center}{\bf \Large Appendix}\end{center}
\vspace{0.08in}

\begin{center}
\begin{minipage}{0.93\linewidth}
\small
\hypersetup{hidelinks}
\hrule
\medskip
\printcontents[appendix]{}{1}[2]{}
\end{minipage}
\end{center}
\vspace{0.08in}

\section{\ourdata construction and characterization}
\label{app:rnadynbench-details}

\subsection{Structure preparation and MD simulation}
\label{app:rnadynbench-simulation}

\Cref{tab:rnadynbench-md-protocol} summarizes the release-level molecular dynamics protocol.
The default solvent box was a rhombic dodecahedron with 12~\AA{} padding.
Systems showing periodic-image contacts during equilibration were rebuilt with 18 or 24~\AA{} padding.
The released files contain the RNA-only post-processed coordinates.

\begin{table}[h]
\centering
\caption{\ourdata molecular dynamics protocol.}
\label{tab:rnadynbench-md-protocol}
\begin{tabular}{p{0.24\linewidth}>{\raggedright\arraybackslash}p{0.68\linewidth}}
    \toprule
    Setting & Value \\
    \midrule
    Simulation engine & GROMACS~\citep{gromacs-2015} \\
    RNA force field & Amber RNA OL3~\citep{AmberRNAOL3-2011} \\
    Solvent and ions & OPC water~\citep{OPCwater-2014}; Li--Merz 12-6 OPC Na$^+$/Cl$^-$ parameters~\citep{Sengupta-2021}; neutralizing ions plus 150 mM NaCl \\
    Periodic box & Rhombic dodecahedron; 12~\AA{} default padding, increased to 18 or 24~\AA{} for periodic-image contacts \\
    Electrostatics & Particle-mesh Ewald~\citep{PME-1993} \\
    Integration & 4 fs time step with hydrogen mass repartitioning~\citep{Feenstra-1999} \\
    Energy minimization & Steepest-descent minimization to the configured force/energy convergence criterion \\
    NVT equilibration & 200 ps at 300 K; Berendsen thermostat~\citep{berendsen1984molecular} \\
    NPT equilibration & 1 ns at 300 K and 1 bar; Nos\'e--Hoover thermostat~\citep{Hose-Hoover-1985} and Berendsen barostat \\
    Final NPT equilibration & 5 ns; Nos\'e--Hoover thermostat and Parrinello--Rahman barostat~\citep{Parrinello-1981} \\
    Production & 100 ns; Nos\'e--Hoover thermostat and Parrinello--Rahman barostat \\
    Coordinate output & 100 ps stride; 1,001 frames from 0 to 100 ns \\
    \bottomrule
    \end{tabular}
\end{table}

\paragraph{Computational resources}
The accepted trajectories were generated on NVIDIA H200, GH200, and A100 GPUs, requiring approximately \textbf{23,000 H200-equivalent GPU hours} including equilibration and production.

\subsection{Quality control and dataset splits}
\label{app:rnadynbench-curation}

We checked residue and atom identities across source structures, prepared RNA systems, GROMACS topologies, and the first released trajectory frames.
Source residue numbers and insertion codes were preserved alongside a sample-level residue index; component-selection and repair decisions were recorded in the metadata.
Structures with missing or ambiguous chain mappings, unresolved internal backbone gaps, unsupported modified residues, or failed topology reconstruction were excluded.

Run-level checks covered energy minimization, equilibration stability, production completion, coordinate spikes, covalent-backbone continuity, and periodic-boundary behavior.
For multi-chain systems, the full 100 ns window was additionally assessed for sustained loss of inter-chain contacts and chain separation.
Release checks verified the monotonic 0--100 ns time axis, 1,001 readable frames, the presence and checksums of all public files, complete metadata, and the absence of exact duplicate trajectory aliases.
Additional geometric and chain-separation findings were recorded as quality-control flags for retained trajectories.

\paragraph{Leakage-controlled splits}
\Cref{tab:rnadynbench-splits} reports the fixed \ourdata partitions and their roles.
The homology-guard trajectories are included in the public release but excluded from training, model selection, and formal evaluation.
The structural test set starts from 113 RNA entries mapped from the gRNAde Das test set~\citep{gRNAde-ICLR2025} and expands by RNA identity and source membership to 117 trajectories.
Corrected source identity is used for grouping while historical RNA-Solo identifiers are retained for provenance.
Direct sequence screening requires at least 80\% identity, defined as the number of matching residues divided by the shorter full RNA sequence length, and at least 80\% coverage on each sequence, measured using actually paired residues.
Multi-chain sequences retain the concatenated representation.
Guards include direct sequence or structural matches and samples sharing a sequence or structural cluster with the structural test set.
Direct structural screening was limited to test--candidate pairs from sequence- or structural-cluster neighborhoods and previously observed direct matches, using a TM-score threshold of 0.45 normalized by the longer RNA.

For \texttt{test\_flex}, each trajectory is scored by its median pairwise aligned C1$'$ RMSD over 101 frames sampled at 1-ns intervals.
After excluding \texttt{test\_struct} and guards, we group trajectories sharing an exact sequence, corrected RNA identity, source structure, sequence cluster, or structural cluster.
Each group receives the median member score.
Eligible groups (at most 25 trajectories) are ranked by score within strata of rounded mean length ($\leq 40$, 41--100, or $>100$ nucleotides), chain multiplicity, predominant source class, and quality-control warnings.
Seeded round-robin selection adds whole groups only when they bring \texttt{test\_flex} closer to 100 trajectories.
A seeded selection from the remaining groups yields 102 validation trajectories; all other groups form the training set.

\begin{table}[h]
    \centering
    \caption{\ourdata dataset splits.}
    \label{tab:rnadynbench-splits}
    \small
    \setlength{\tabcolsep}{5pt}
    \begin{tabular}{@{}lrl@{}}
        \toprule
        Partition & Trajectories & Primary role \\
        \midrule
        Training & 2,248 & Model optimization and downstream probe training \\
        Validation & 102 & Checkpoint and hyperparameter selection \\
        \texttt{test\_struct} & 117 & Structural and molecular generalization \\
        \texttt{test\_flex} & 100 & High-flexibility generalization \\
        Homology guard & 18 & Excluded homology/cluster guards \\
        \midrule
        Total & 2,585 & \\
        \bottomrule
    \end{tabular}
\end{table}

\subsection{Size-matched RNA--protein comparisons}
\label{app:rnadynbench-characterization}

For each protein reference, we retained the overlapping heavy-atom-count range and partitioned log heavy-atom count into five bins.
RNA and protein systems then received common bin weights.
ATLAS and dynamicPDB were matched separately because their size distributions differed.

\begin{figure}[H]
    \centering
    \includegraphics[width=\textwidth]{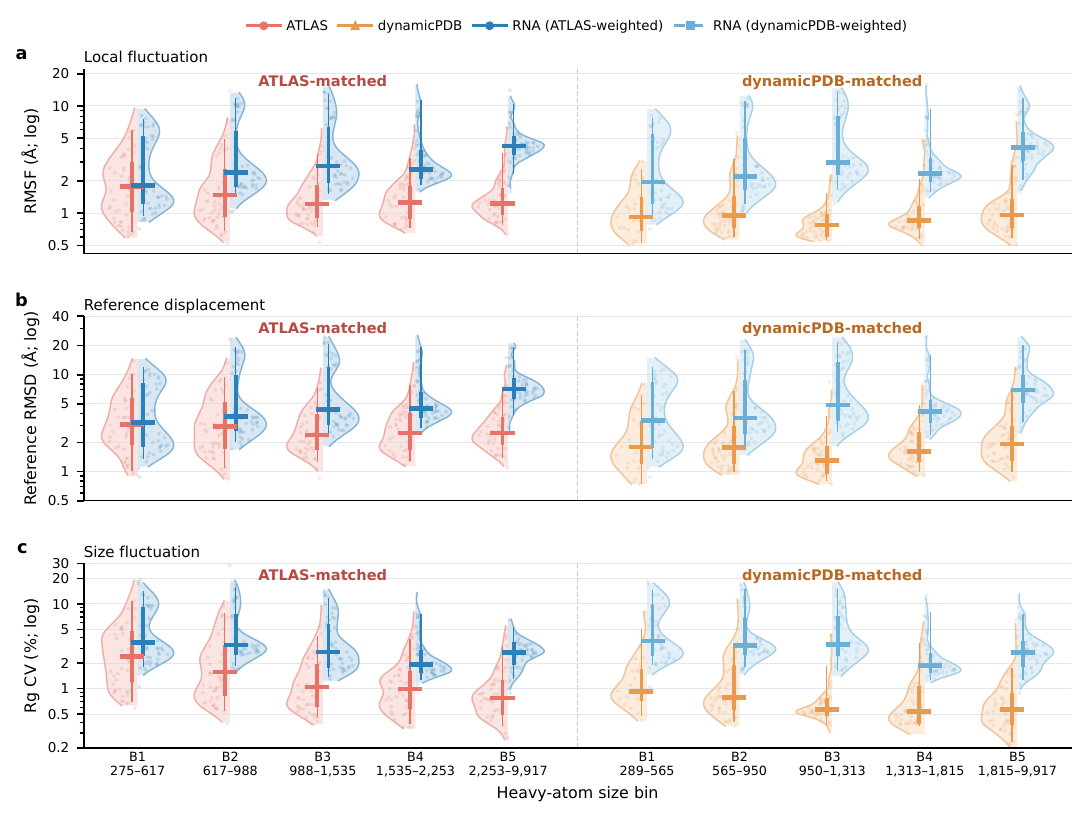}
    \caption{\textbf{Size-stratified RNA--protein dynamics.}
Distributions of (\textbf{a}) mean marker-atom RMSF, (\textbf{b}) mean RMSD to the first sampled frame, and (\textbf{c}) heavy-atom $R_\mathrm{g}$ coefficient of variation across five size bins.
RNA uses C4$'$ markers and protein uses C$\alpha$; vertical axes are logarithmic.
Split violins show proteins on the left and matched RNAs on the right; bars indicate the median, interquartile range, and 5th--95th percentiles.}
    \label{fig:rna-protein-stratified}
\end{figure}

\section{\ourmodel architecture and objectives}
\label{app:model-details}

Bold symbols denote vectors, matrices, or tensors.
Subscript $f$ indexes frames, and $i,j$ index residues.
\Cref{tab:notation} summarizes the principal notation for \ourmodel.
Quantities used only in the downstream binding model are defined in its own section.

\begin{table}[htbp]
\centering
\caption{Notation used throughout our manuscript.}
\label{tab:notation}
\small
\setlength{\tabcolsep}{4pt}
\renewcommand{\arraystretch}{1.12}
\begin{tabular}{@{}p{0.31\linewidth}p{0.65\linewidth}@{}}
\toprule
Symbol & Meaning \\
\midrule

$L,A,N,d$
& Numbers of frames, atoms, residues, and feature dimensions. \\

$\mathbf X,\mathbf x_f$
& All-atom trajectory window and its frame-$f$ conformer. \\

$\theta,\bar\theta$
& Online parameters and their EMA; fingerprints use the latter. \\

$\mathcal D_\theta,\mathcal E_\theta,\mathcal E_{\bar\theta},\ell_{\mathrm{fp}}$
& Trajectory denoiser, online single-conformer encoder, EMA fingerprint encoder, and readout block. \\

$\widetilde{\mathbf X},\boldsymbol{\sigma},\boldsymbol\epsilon$
& Noised trajectory, coordinate-noise scale tensor, and Gaussian perturbation. \\

$\mathbf R_{\mathrm{traj}},\mathbf R_{\mathrm{sf}}$
& Independent global rotations of the trajectory and single-frame views. \\

$J,\mathcal V$
& Sampled anchor-frame index and set of valid residues. \\

$\mathbf H_{\bar\theta}^{\mathrm{FP}},\mathbf h_{\bar\theta,i}^{\mathrm{FP}}$
& EMA single-conformer fingerprint matrix and its residue-$i$ vector. \\

$\mathbf h_{\theta,i}^{\mathrm{FP}},\mathbf h_{\theta,i}^{\mathrm{traj}}$
& Online single-conformer and trajectory-anchor states for residue $i$ at sampled anchor $J$. \\

$q_\psi,\mathbf z_i^{\mathrm{traj}}$
& Alignment predictor conditioned on $H$ and $\Delta t$, and detached trajectory-anchor target. \\

$\boldsymbol\Sigma_i,\mathbf R_i^{(J)}$
& Residue covariance and orthonormal residue frame of clean anchor $J$. \\

$\mathbf C,\boldsymbol\rho$
& Inter-residue motion-coupling matrix and its normalized form (NMC). \\

$f_\Sigma,f_C$
& Covariance and motion-coupling heads for physical grounding. \\

$\mathcal L_{\mathrm{gen}},\mathcal L_{\mathrm{align}},\mathcal L_{\mathrm{PG}}$
& Generation, single-frame-to-trajectory alignment, and physical-grounding objectives. \\

\bottomrule
\end{tabular}
\end{table}

\subsection{Backbone and temporal conditioning}
\ourmodel initializes its diffusion backbone from Protenix \citep{protenix-2025}.
Following Protenix's AF3-style tokenization, standard RNA nucleotides form residue tokens, and the frozen Input Embedder and Pairformer in \cref{fig:framework} provide single and pair features.
An atom encoder maps EDM-preconditioned noisy coordinates to residue tokens before spatiotemporal attention.
Each spatiotemporal block applies spatial attention followed by temporal self-attention.
The diffusion backbone is optimized jointly with the representation heads.
The latent objectives use an intermediate token representation, whereas coordinate prediction continues through the denoising path.

\paragraph{Time conditioning and the single-frame predictor}

The alignment target of Eq.~\eqref{eq:trajectory-target} depends on the horizon $H$ and interval $\Delta t$ of the window, so the single-frame predictor is conditioned on both.
The predictor embeds standardized logarithms of $H$ and $\Delta t$:
\begin{equation}
    \mathbf e_{\mathrm{time}}
    =
    \operatorname{MLP}_{\mathrm{time}}
    \left(
        \left[
            \frac{\log H-\mu_H}{s_H},
            \frac{\log\Delta t-\mu_{\Delta t}}{s_{\Delta t}}
        \right]
    \right),
    \label{eq:time-embedding}
\end{equation}
and the shared residue-wise predictor applies FiLM modulation,
\begin{equation}
\begin{aligned}
    \left[
        \boldsymbol{\gamma}_{\mathrm{time}},
        \boldsymbol{\beta}_{\mathrm{time}}
    \right]
    &=
    W_{\mathrm{FiLM}}
    \mathbf e_{\mathrm{time}},\\
    q_{\psi}
    \left(
        \mathbf h,H,\Delta t
    \right)
    &=
    \mathbf W_2
    \operatorname{GELU}
    \left[
        \left(
            1+\boldsymbol{\gamma}_{\mathrm{time}}
        \right)
        \odot
        \left(
            \mathbf W_1\mathbf h+\mathbf b_1
        \right)
        +
        \boldsymbol{\beta}_{\mathrm{time}}
    \right]
    +
    \mathbf b_2.
\end{aligned}
    \label{eq:time-predictor}
\end{equation}
The FiLM projection is zero-initialized, so the predictor initially applies no time-dependent modulation.
The statistics $(\mu_H,s_H,\mu_{\Delta t},s_{\Delta t})$ are computed once from the training-window distribution and held fixed during training.

\subsection{Training views and fingerprint extraction}
\label{app:training-views}

Each sampled trajectory is rigidly aligned to its first frame and
centered before view construction.
Let $\mathcal V$ denote the RNA residues with at least one resolved atom.

\paragraph{View construction}
The trajectory view uses one global rotation
$\mathbf R_{\mathrm{traj}}$ across all frames, applied jointly to
coordinates and Gaussian noise.
Its reconstruction target is $\mathbf R_{\mathrm{traj}}\mathbf X$.
The single-frame pass receives the corresponding clean anchor
$\mathbf R_{\mathrm{sf}}\mathbf x_J$, without coordinate noise.
The two rotations are sampled independently and uniformly from
$\mathrm{SO}(3)$.

\paragraph{Readouts and alignment target}
Let $\mathcal H_\theta^{\ell_{\mathrm{fp}}}$ denote the online backbone
states at the fingerprint readout block.
For the sampled anchor $J$, the trajectory readout and alignment target are
\begin{equation}
    \begin{aligned}
        \mathbf h_{\theta,i}^{\mathrm{traj}}
        &=\operatorname{LN}\!\left[
            \mathcal H_\theta^{\ell_{\mathrm{fp}}}
            \left(
                \widetilde{\mathbf X},
                \boldsymbol{\sigma},
                H,\Delta t
            \right)_{J,i}
        \right],\\
        \mathbf z_i^{\mathrm{traj}}
        &=\operatorname{sg}\!\left[
            \operatorname{norm}
            \left(\mathbf h_{\theta,i}^{\mathrm{traj}}\right)
        \right].
    \end{aligned}
    \label{eq:trajectory-target}
\end{equation}
Here $\operatorname{LN}$ is affine-free LayerNorm,
$\operatorname{norm}(\mathbf z)
=\mathbf z/(\|\mathbf z\|_2+\varepsilon)$,
and $\operatorname{sg}$ stops gradients.
The normalized, detached target is read at one anchor after temporal attention over the full window.
The single-frame readout uses affine-free LayerNorm at the same depth, before the predictor.
\Cref{eq:alignment-loss} averages the alignment loss over valid residues; its gradients update the single-frame path.

Reported single-conformer fingerprints use the EMA encoder in evaluation mode on one centered conformer, without random rotation or coordinate noise.
Any probe-specific normalization is applied after this readout.

\subsection{Trajectory generation objective}
\label{app:training-losses}

The generation objective combines coordinate denoising, local structural agreement, fluctuation matching, and short-lag autocorrelation losses:
\begin{equation}
\begin{aligned}
    \mathcal L_{\mathrm{gen}}
    ={}&\mathcal L_{\mathrm{denoise}}
    +\lambda_{\mathrm{sLD}}\mathcal L_{\mathrm{smooth\text{-}lDDT}}
    +\lambda_{\mathrm{RMSF}}\mathcal L_{\mathrm{RMSF}}\\
    &+\lambda_{\mathrm{rel}}\mathcal L_{\mathrm{rel\text{-}RMSF}}
    +\lambda_{\mathrm{local}}\mathcal L_{\mathrm{local\text{-}RMSF}}
    +\lambda_{\mathrm{ACF}}\mathcal L_{\mathrm{ACF}}.
\end{aligned}
\label{eq:generation-loss}
\end{equation}
For the reported model, the effective multipliers are
$\lambda_{\mathrm{sLD}}=4$, $\lambda_{\mathrm{RMSF}}=0.4$,
$\lambda_{\mathrm{rel}}=\lambda_{\mathrm{local}}=\lambda_{\mathrm{ACF}}=3.2$.
The denoising loss in \cref{eq:denoise} already includes its generative-group multiplier.

The coordinate objective follows BioKinema's global-frame reconstruction convention, without fitting a new rigid alignment between the predicted and target coordinates.
For each sampled frame, squared Euclidean errors are weighted by atom validity, molecular type, and the residue-specific EDM factor.
Weighted errors are summed over atoms, divided by the number of resolved atoms and by three spatial dimensions, and averaged over all sampled frames.
Thus the coordinate MSE includes near-clean conditioning frames; AF3-style smooth-lDDT \citep{AF3-Nature2024} instead uses the explicit target-frame mask to supervise local structural agreement.
The RMSF-based terms follow BioKinema's fluctuation supervision \citep{BioKinema-bioRxiv2026}.
The RMSF term compares per-atom positional fluctuations over the window.
Relative and local RMSF compare fluctuations of selected inter-token and within-token atom-pair distances, respectively.
The ACF term compares normalized, time-centered positional autocorrelations over short frame lags for selected atoms.
The RMSF, relative-RMSF, local-RMSF, and ACF terms apply only to windows with at least three frames.

\subsection{Physical grounding targets and prediction heads}
\label{app:physical-grounding}

\paragraph{Reference statistics and local coordinates}
PG uses the same 101-frame, 100-ns reference trajectory for covariance and motion-coupling labels, with population normalization over all 101 frames.
We align all frames to frame zero by a C1$'$-based rigid fit and remove the temporal mean separately for each residue.
The labels satisfy $C_{ii}=\operatorname{tr}(\boldsymbol\Sigma_i)$, while $\mathbf C$ is invariant to a common global rotation.
For each clean anchor, the first local axis points from C1$'$ to C3$'$; the second is the C1$'$-to-C4$'$ vector orthogonalized against the first, and the third is their cross product.
The three normalized axes form the columns of $\mathbf R_i^{(J)}$.
Full-trajectory statistics are expressed in this clean-anchor frame, even when the supervised hidden state comes from the noisy trajectory pass.
Labels are matched by residue identity and only valid residues enter the reductions.

The main text uses numerical covariance values in \AA$^2$.
With $s_0=1\,\text{\AA}^2$, the matrix logarithm acts on the dimensionless local covariance $\boldsymbol\Sigma_{i,\mathrm{loc}}^{(J)}/s_0$ after eigenvalue flooring.
The physical prediction is recovered as $\widehat{\boldsymbol\Sigma}_i=s_0\mathbf R_i\exp\!\bigl(f_\Sigma(\mathbf h_{\bar\theta,i}^{\mathrm{FP}})\bigr)\mathbf R_i^{\top}$.

\paragraph{Shared covariance head}
The single-conformer and trajectory-anchor representations are read at the same intermediate block.
The covariance head is a normalized multilayer perceptron with six outputs.
Its six outputs fill the independent entries of a symmetric $3\times3$ matrix $\mathbf S$, directly parameterizing the log-covariance.
Training compares $\mathbf S$ with detached matrix-log targets, without differentiating through an eigendecomposition of the prediction.
The matrix-entry loss in \cref{eq:pg-covariance} averages squared errors over all nine entries, valid residues, and the two views.
Each symmetric off-diagonal entry contributes twice.
The matrix exponential recovers covariance descriptors at inference and is absent from the training loss.

\paragraph{Motion-coupling head and reductions}
The motion-coupling head reads intermediate residue states.
A normalized residue-level multilayer perceptron produces vectors $\mathbf u_i$; symmetric pair features $[\mathbf u_i+\mathbf u_j;|\mathbf u_i-\mathbf u_j|]$ are mapped to a scalar matrix $\mathbf A_C$.
Its lower-triangular factor is
\begin{equation}
    \mathbf L_C=\operatorname{tril}(\mathbf A_C,-1)
    +\operatorname{diag}\!\left(\operatorname{softplus}(\operatorname{diag}\mathbf A_C)+\varepsilon_C\right),
    \label{eq:coupling-factor}
\end{equation}
where $\varepsilon_C>0$ stabilizes the diagonal and $\widehat{\mathbf C}=\mathbf L_C\mathbf L_C^\top$ is positive definite.
The lower-triangular factor is constructed in the input residue order. The readout applies an MLP to \ourmodel's residue states in place of BackFlip's IPA edge encoder.

In \cref{eq:coupling-loss}, the matrix-pattern and magnitude terms include diagonal entries, while NMC MSE uses off-diagonal entries only.
Small positive floors stabilize the Frobenius-norm denominator in \cref{eq:coupling-loss} and NMC normalization.
NMC is derived from the same motion-coupling matrix, with no separate prediction head.
The per-sample losses are averaged over the batch.
The covariance and coupling heads are supervised separately, without an explicit cross-head consistency constraint.
PG supervises covariance and zero-lag motion coupling; the generation objective separately supervises short-lag autocorrelation.
The combined objective is $\mathcal L_{\mathrm{PG}}=\lambda_\Sigma\mathcal L_\Sigma+\lambda_C\mathcal L_C$.

\section{Experimental protocols}
\label{app:experimental-setup}

The code and configurations specify remaining masks and reductions, head dimensions, numerical constants, objective weights, and training, sampling, and downstream optimization settings.

\subsection{Training configuration}

Training uses the official \ourdata training partition, with variable window starts, spacings, and lengths.
Stored-frame spacing is converted to the physical interval $\Delta t$ for temporal conditioning.
The sampler mixes unconditional, initial-frame-conditioned, and endpoint-conditioned windows while varying their temporal extent and resolution.

Training starts from the Protenix \citep{protenix-2025} initialization with a fresh
optimizer state.
We use validation-guided optimization, learning-rate scheduling,
gradient clipping, mixed-precision training, and an EMA copy for
trajectory generation.
PG is active from the first update.
Single-frame-to-trajectory alignment is activated after an initial warm-up and
smoothly increased to its final weight, using one sampled anchor per
eligible forward window.
The alignment target is obtained from the online trajectory pass, while
EMA weights are retained for generation.

\paragraph{Evaluation sets}
\ourdata defines \texttt{test\_struct} as 117 trajectories from 62 RNAs and \texttt{test\_flex} as 100 trajectories from 65 RNAs. Generation results for \ourmodel{} and frozen-probe evaluations for \ourmodel{}, w/o PG, and w/o alignment and PG use these panels.
BioKinema and \ourmodel generation use fixed EMA checkpoints.

\subsection{Trajectory sampling}
\label{app:trajectory-sampling}

Both test panels use the same EMA denoiser and hierarchical forecasting-and-interpolation protocol.
The coarse stage generates temporally spaced anchors from the supplied initial conformer.
The fine stage uses the denoiser conditioned on adjacent anchors to generate intermediate frames on the 101-frame, 100-ns evaluation grid.
Each of the four samples is a complete trajectory.

\subsection{Baselines and ablation settings}
\label{app:baseline-details}

\paragraph{Trajectory-generation baselines}
The baselines use distinct RNA input representations.
BioKinema uses precomputed Protenix single and pair features for all-atom coordinate denoising and is retrained on \ourdata.
Our MDGen adaptation is trained from scratch and retains the original flow-matching architecture and sampler.
It represents each nucleotide by a rigid-frame offset and ten sine--cosine torsion pairs (27 dimensions), with RNA-specific residue identities and coordinate decoding.
Validation-selected EMA weights are used for evaluation.
At evaluation, decoded frame zero is replaced by the input conformer; generated future frames are left unchanged.
ConfRover uses RNA frames and ten torsion pairs, C1$'$ pair geometry, and frozen Protenix features from sequence and topology.
Its causal temporal and SE(3) diffusion modules and RNA torsion head are trained from scratch; an RNA decoder reconstructs atomic coordinates.
ConfRover generates one trajectory per target ($K=1$); its runtime is per trajectory, and it is excluded from the $K=4$ ranking.

\paragraph{Descriptor baselines}
All descriptor models receive the experimental initial structure.
Geometry-GNN builds a distance-based graph from a single RNA conformer and jointly trains its graph network and descriptor heads on MD-derived targets.
RNA-FM+2D GCN freezes RNA-FM but trains its GCN and prediction heads.
Other unmarked model rows use native descriptor heads.
DynaProt combines validation-selected marginal and joint native heads.
Rows marked with $\dagger$ use independently trained descriptor probes on frozen representations.

\paragraph{Representation-learning ablations}
The \textit{w/o align.\ \& PG} variant uses only $\mathcal L_{\mathrm{gen}}$; \textit{w/o PG} adds $\mathcal L_{\mathrm{align}}$, and the full model also includes PG.
All variants retain the coordinate, structural, fluctuation, and short-lag correlation losses in $\mathcal L_{\mathrm{gen}}$ (\cref{app:training-losses}).
For their $\dagger$ rows, we freeze the EMA encoders and train separate probes on MD-derived descriptors.

\subsection{Evaluation metrics and MD references}
\label{app:evaluation-metrics}

\paragraph{Trajectory generation}
Geometry metrics use the 100 future frames; frame zero is the supplied initial conformer.
Bond MAE averages absolute RNA heavy-atom bond-length deviations from that conformer.
O3$'$--P max is the largest absolute initial-linkage deviation over future frames and consecutive within-chain residues.
The maximum is computed per trajectory before averaging generated samples.
Chiral flip is the percentage of sugar-stereocenter observations with a signed-volume sign different from frame zero, including zero volume.
Clashes count nonbonded heavy-atom pairs closer than 1.5~\AA{}, excluding directly bonded pairs and pairs sharing a bonded neighbor.
Counts are normalized per 1,000 heavy atoms.
For dynamics metrics, C1$'$ coordinates in every frame are rigidly aligned to the MD initial conformer without reflection.
RMSF $r$ is the Pearson correlation across residues between MD and the generated-sample-mean RMSF profiles, each computed over all 101 frames.
PCA-$W_2$ compares equal-weight distributions of the 100 future frames in a two-dimensional PCA space fitted only to MD.
It uses optimal assignment and normalization by the square root of the C1$'$ atom count.
Framewise geometry metrics are averaged over frames and generated samples, then within RNA systems and across RNAs.
Per-target RMSF correlations and sample-mean PCA-$W_2$ values are likewise averaged within RNAs and then across RNAs.

\paragraph{Single-conformer dynamics prediction}
Model predictions are compared with descriptors from the full 100-ns MD trajectory.
The MD reference instead compares two non-overlapping 50-ns blocks from that same trajectory to indicate finite-sampling variability.
As in prior covariance-only evaluations \citep{DynaProt-ICLR2026,BackFlip2-arXiv2026}, we compare predicted and MD residue fluctuations as zero-mean Gaussians.
Both covariance matrices are symmetrized and projected to positive definiteness before computing RMWD, symmetric KL, and IoU; the equations below use these projected matrices.
For $N$ evaluated residues, root-mean 2-Wasserstein distance is
\begin{equation}
\mathrm{RMWD}=\left[\frac{1}{N}\sum_{i=1}^{N}\operatorname{tr}\left(
\widehat{\boldsymbol\Sigma}_i+\boldsymbol\Sigma_i
-2\left(\boldsymbol\Sigma_i^{1/2}\widehat{\boldsymbol\Sigma}_i\boldsymbol\Sigma_i^{1/2}\right)^{1/2}
\right)\right]^{1/2}.
\label{eq:evaluation-rmwd}
\end{equation}
The symmetric Kullback--Leibler divergence is
\begin{equation}
\mathrm{KL}_{\mathrm{sym}}=\frac{1}{4N}\sum_{i=1}^{N}\left[
\operatorname{tr}(\boldsymbol\Sigma_i^{-1}\widehat{\boldsymbol\Sigma}_i)
+\operatorname{tr}(\widehat{\boldsymbol\Sigma}_i^{-1}\boldsymbol\Sigma_i)-2d\right],\qquad d=3.
\label{eq:evaluation-symmetric-kl}
\end{equation}
Because both Gaussians are centered, neither metric includes mean-coordinate error.
Ellipsoid IoU also evaluates the shape and orientation of the covariance:
\begin{equation}
\mathcal E(\boldsymbol\Sigma_i)
=\{\mathbf x\in\mathbb R^3:
\mathbf x^{\top}\boldsymbol\Sigma_i^{-1}\mathbf x\le 1\}.
\label{eq:covariance-ellipsoid}
\end{equation}
Its eigenvectors define principal directions, and square-root eigenvalues define semi-axis lengths.
IoU averages each residue's ellipsoid-intersection volume divided by its union volume, estimated by spherical quadrature.
Thus, IoU reflects fluctuation scale as well as anisotropy and orientation.

RMSF and NMC are derived as in \cref{eq:pg-targets}.
For RMSF, Pearson $r$ compares residue-wise flexibility patterns, while MAE measures absolute fluctuation-amplitude error.
For NMC, $r$ compares coupling patterns and MAE measures absolute coupling error over strictly upper-triangular residue pairs ($i<j$).
RMSF uses the raw covariance trace rather than the projected matrices used for RMWD, KL, and IoU.
Metrics are computed per trajectory, averaged within each RNA group, and reported as split-level RNA-group macro means.
RMWD and RMSF MAE are in \AA{}; the other descriptor scores are dimensionless.

\paragraph{Generation reference and aggregation}
BioKinema, MDGen, and \ourmodel generate four trajectories per target.
Reference-MD RNA-level intervals use cluster resampling.
Geometry reference values are empirical RNA-level means from MD.
Trajectory-matching reference values are medians of RNA-level statistics from conditional moving-block bootstrap samples of individual MD trajectories.
These intervals describe resampling variability conditional on the observed trajectories.

\subsection{RNA--ligand binding protocol}
\label{app:binding}

This subsection details the binding pipeline summarized in \cref{sec:exp-rna-ligand-binding}.
RNA residues are represented by combining the frozen residue-level fingerprint $\mathbf h_{\bar\theta,i}^{\mathrm{FP}}$ exported by \ourmodel with frozen sequence and secondary-structure features:
\begin{equation}
        \widetilde{\mathbf h}_i^{\mathrm{FP}}
        =
        \operatorname{MLP}_{\mathrm{FP}}
        \left(\mathbf h_{\bar\theta,i}^{\mathrm{FP}}\right),
        \quad
        \mathbf h_{\mathrm{RNA},i}
        =
        \operatorname{MLP}_{\mathrm{fuse}}
        \left(
            \widetilde{\mathbf h}_i^{\mathrm{FP}}
            \mathbin\Vert
            \mathbf e_i^{1\mathrm{D}}
            \mathbin\Vert
            \mathbf e_i^{2\mathrm{D}}
        \right).
    \label{eq:binding-entity-encoding}
\end{equation}

The ligand encoder combines atom representations from a 2D graph convolutional network and a 3D Equiformer.
The GerNA-Bind interaction stack~\citep{GerNABind-NMI2025} constructs residue--atom pair features, refines them with triangle multiplication and row-wise attention using intramolecular RNA and ligand distances, and pools the pair tensor with dual mutual attention.
A prediction head maps the pooled representation to a binding score.
Training uses the binary evidential objective from the GerNA-Bind implementation.
No RNA--ligand complex pose or cross-molecule distance is required.
Downstream optimization follows the released implementation.
The fingerprint encoder and the pretrained RNA and ligand encoders from GerNA-Bind were frozen.
The new RNA and ligand fusion layers, the retained interaction stack, and the prediction head were optimized from the released initialization.
This setup evaluates the downstream utility of frozen \ourmodel fingerprints within the GerNA-Bind interaction and prediction framework.

\section{Supplementary results}
\label{app:additional-results}

\subsection{Grounding targets and supervision depth}
\label{app:ablations}

\Cref{tab:generation-ablation,tab:prediction-ablation} compare the full model with variants retaining only covariance or only motion-coupling grounding.
Block 24 moves the fingerprint readout and its alignment and PG supervision from Block 17 to Block 24; Block 17 supplies the main-text results.
All generation variants use the same coarse and fine sampling schedule.
Block 17 has the highest generation RMSF correlation on both test panels, whereas Block 24 has lower predictive RMWD on both panels.
\begin{table}[H]
    \centering
    \caption{Trajectory-generation ablations. RNA-group macro means; distances are in \AA{} and chiral flips in percent.}
    \label{tab:generation-ablation}
    \scriptsize
    \setlength{\tabcolsep}{2pt}
    \resizebox{\linewidth}{!}{%
    \begin{tabular}{@{}lrrrrrr@{\hspace{4pt}}rrrrrr@{}}
        \toprule
        \multirow{2}{*}[-2.6ex]{Model} & \multicolumn{6}{c}{\texttt{test\_struct}} & \multicolumn{6}{c}{\texttt{test\_flex}} \\
        \cmidrule(lr){2-7}\cmidrule(l){8-13}
         & \makecell{Bond\\MAE$\downarrow$} & \makecell{O3P\\max$\downarrow$} & \makecell{Chiral\\flip$\downarrow$} & \makecell{Clash\\/1k$\downarrow$} & \makecell{RMSF\\$r\uparrow$} & \makecell{PCA\\$W_2\downarrow$} & \makecell{Bond\\MAE$\downarrow$} & \makecell{O3P\\max$\downarrow$} & \makecell{Chiral\\flip$\downarrow$} & \makecell{Clash\\/1k$\downarrow$} & \makecell{RMSF\\$r\uparrow$} & \makecell{PCA\\$W_2\downarrow$} \\
        \midrule
        Cov-only & 0.02423 & 1.7091 & 0.03092 & 0.9583 & 0.8712 & 1.8699 & 0.02561 & 3.8573 & 0.00126 & 3.2085 & 0.7609 & 5.8609 \\
        Coupling-only & 0.02414 & 1.0845 & 0.03088 & 0.9891 & 0.8631 & 1.8983 & 0.02537 & 5.4953 & 0.00093 & 2.8008 & 0.7623 & 5.8490 \\
        Full model (Block 24) & 0.02430 & 1.2558 & 0.03091 & 0.9256 & 0.8720 & 1.8896 & 0.02586 & 6.2126 & 0.00043 & 3.1947 & 0.7592 & 5.8033 \\
        \midrule
        Full model (Block 17) & 0.02424 & 1.8858 & 0.03085 & 0.8225 & 0.8746 & 1.8739 & 0.02567 & 4.2400 & 0.00093 & 3.0404 & 0.7658 & 5.9169 \\
        \bottomrule
    \end{tabular}%
    }
\end{table}

\begin{table}[H]
    \centering
    \caption{Single-conformer dynamics ablations using independent probes on frozen EMA backbones. RNA-group macro means; RMWD and RMSF MAE are in \AA{}.}
    \label{tab:prediction-ablation}
    \scriptsize
    \setlength{\tabcolsep}{2pt}
    \resizebox{\linewidth}{!}{%
    \begin{tabular}{@{}lrrrrrrr@{\hspace{1pt}}rrrrrrr@{}}
        \toprule
        & \multicolumn{7}{c}{\texttt{test\_struct}} & \multicolumn{7}{c}{\texttt{test\_flex}} \\
        \cmidrule(lr){2-8}\cmidrule(l){9-15}
        & \multicolumn{2}{c}{RMSF} & \multicolumn{3}{c}{Covariance} & \multicolumn{2}{c}{NMC}
        & \multicolumn{2}{c}{RMSF} & \multicolumn{3}{c}{Covariance} & \multicolumn{2}{c}{NMC} \\
        \cmidrule(lr){2-3}\cmidrule(lr){4-6}\cmidrule(lr){7-8}\cmidrule(lr){9-10}\cmidrule(lr){11-13}\cmidrule(l){14-15}
        Method & $r\uparrow$ & MAE$\downarrow$ & RMWD$\downarrow$ & Sym. KL$\downarrow$ & IoU$\uparrow$ & $r\uparrow$ & MAE$\downarrow$ & $r\uparrow$ & MAE$\downarrow$ & RMWD$\downarrow$ & Sym. KL$\downarrow$ & IoU$\uparrow$ & $r\uparrow$ & MAE$\downarrow$ \\
        \midrule
        Cov-only & 0.8540 & 0.4307 & 0.8856 & 0.6007 & 0.5309 & 0.7936 & 0.1722 & 0.7711 & 1.7128 & 3.3259 & 0.8238 & 0.4547 & 0.7537 & 0.2090 \\
        Coupling-only & 0.8448 & 0.4535 & 0.9297 & 0.6690 & 0.4806 & 0.8160 & 0.1626 & 0.7468 & 1.7138 & 3.3845 & 0.9596 & 0.4322 & 0.7646 & 0.2012 \\
        Full model (Block 24) & 0.8622 & 0.4124 & 0.8716 & 0.5892 & 0.5276 & 0.8016 & 0.1659 & 0.7696 & 1.5995 & 3.2786 & 0.8498 & 0.4476 & 0.7592 & 0.2023 \\
        \midrule
        Full model (Block 17) & 0.8626 & 0.4076 & 0.8741 & 0.5823 & 0.5276 & 0.7958 & 0.1694 & 0.7782 & 1.7560 & 3.3359 & 0.8775 & 0.4522 & 0.7577 & 0.2035 \\
        \bottomrule
    \end{tabular}%
    }
\end{table}

\clearpage
\subsection{Predictive data scaling}
\begin{figure}[!htbp]
    \centering
    \includegraphics[width=\textwidth]{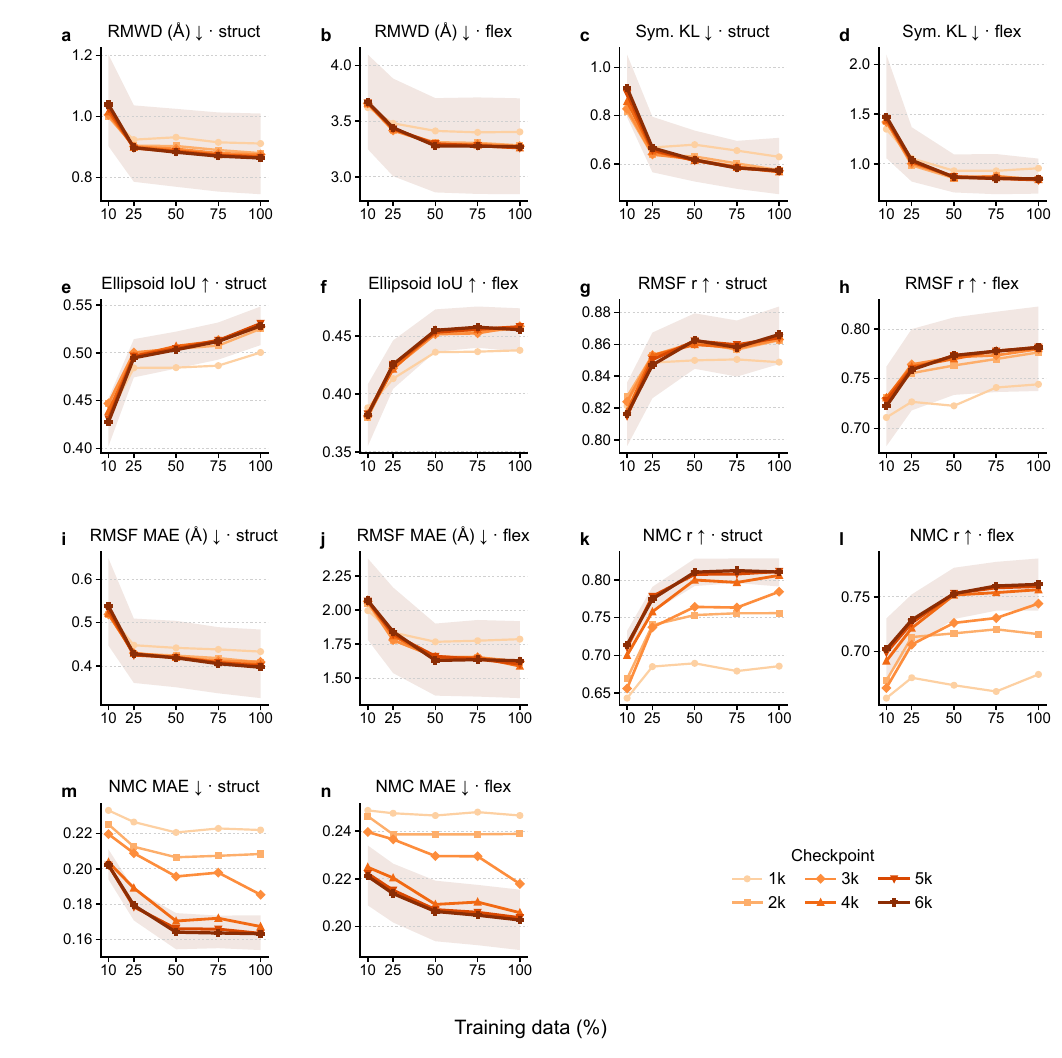}
    \caption{\textbf{Predictive data scaling.}
    Orange curves darken from 1k to 6k training updates across 10--100\% of the training data.
    Shading shows pointwise 95\% source-RNA-group bootstrap intervals for 6k only.
    Each checkpoint uses its EMA backbone and EMA descriptor heads.}
    \label{fig:predictive-scaling-full}
\end{figure}

\Cref{fig:predictive-scaling-full} shows the seven metrics reported in
\cref{tab:dynamics} across both held-out splits and all
training-data fractions and checkpoints.

\subsection{Qualitative trajectory generation}
\label{app:qualitative-generation}

\begin{figure}[p]
    \centering
    \includegraphics[width=\textwidth]{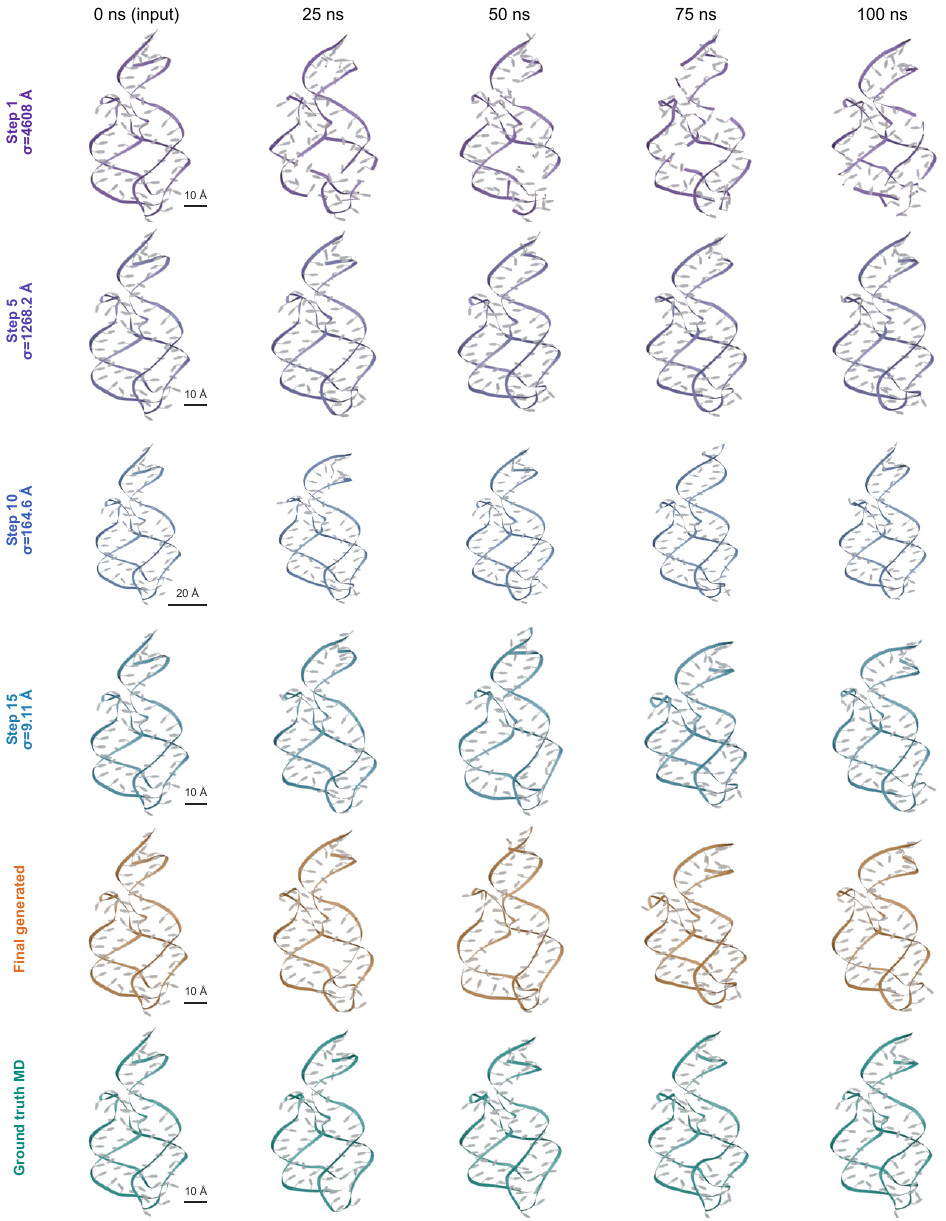}
    \caption{\textbf{Stepwise denoising of the low-motion held-out RNA 3IVN\_A.}
    Columns show 0, 25, 50, 75, and 100 ns. Step rows are diagnostic $\hat{\mathbf x}_0$ estimates at the labeled noise scales; the final two rows show the generated trajectory and reference MD.
    Ribbon gaps mark O3$'$--P distances of at least 2.6~\AA{}.}
    \label{fig:denoising-low}
\end{figure}

\begin{figure}[p]
    \centering
    \includegraphics[width=\textwidth]{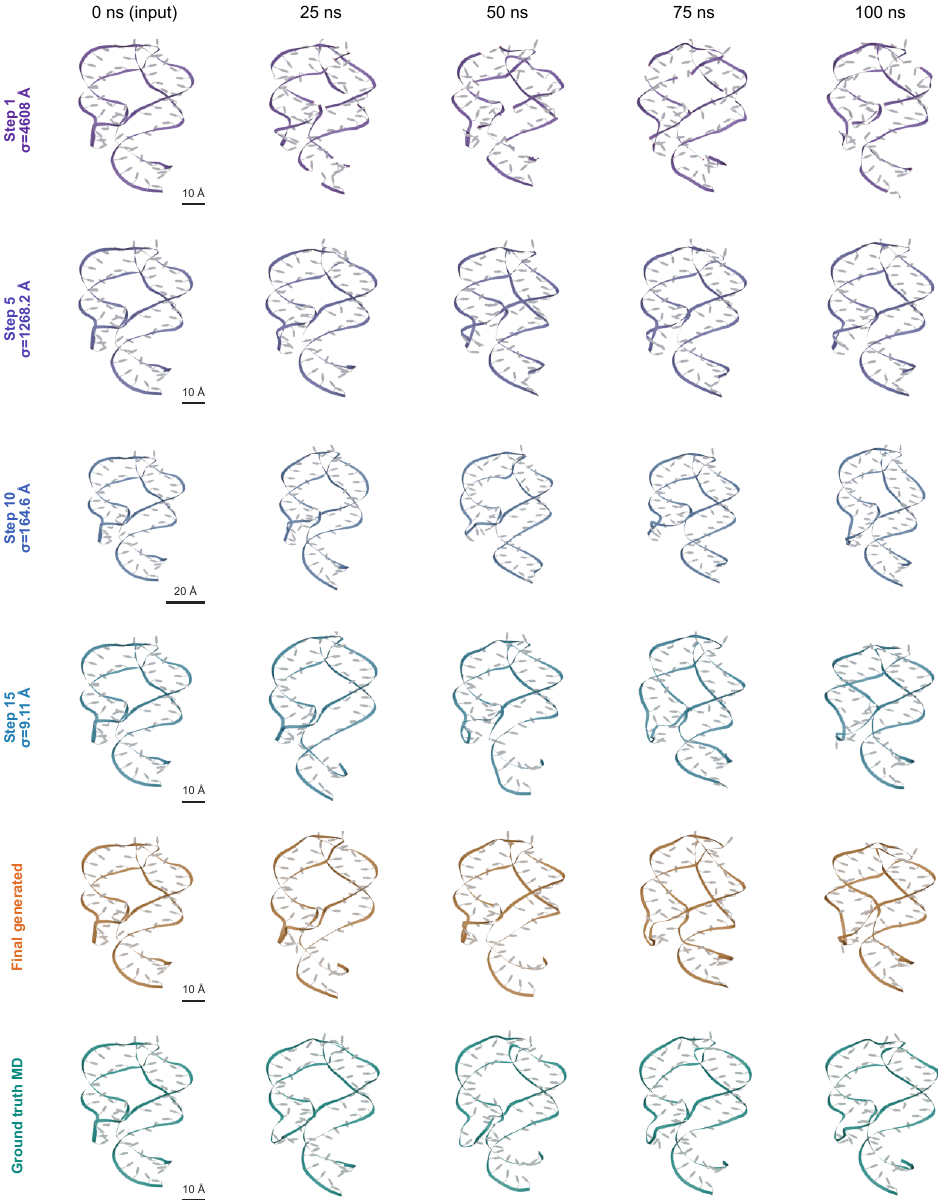}
    \caption{\textbf{Stepwise denoising of the median-motion held-out RNA 3SKR\_A.}
    Columns show 0, 25, 50, 75, and 100 ns. Step rows are diagnostic $\hat{\mathbf x}_0$ estimates at the labeled noise scales; the final two rows show the generated trajectory and reference MD.
    Ribbon gaps mark O3$'$--P distances of at least 2.6~\AA{}.}
    \label{fig:denoising-medium}
\end{figure}

\begin{figure}[p]
    \centering
    \includegraphics[width=\textwidth]{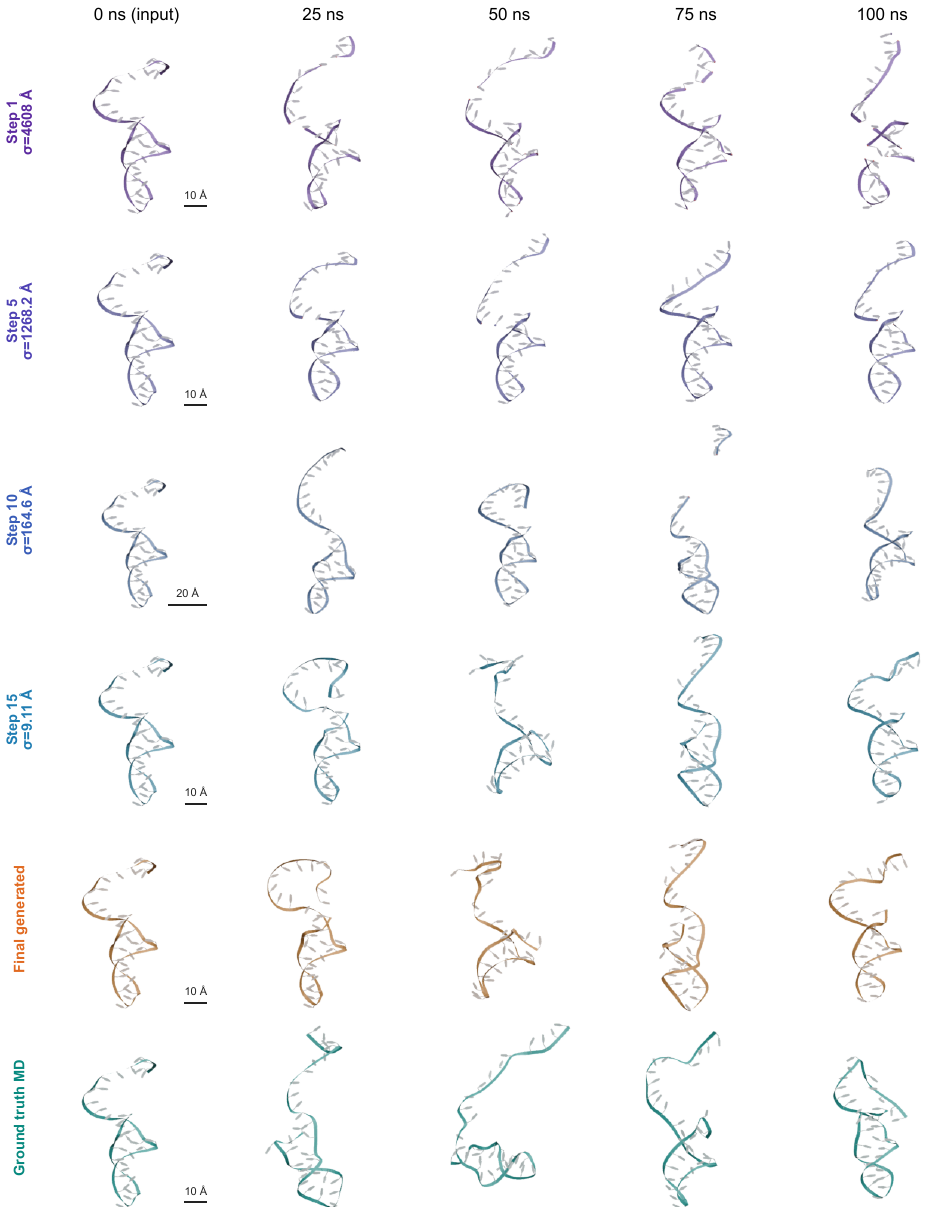}
    \caption{\textbf{Stepwise denoising of the high-motion held-out RNA 6AZ4\_A.}
    Columns show 0, 25, 50, 75, and 100 ns. Step rows are diagnostic $\hat{\mathbf x}_0$ estimates at the labeled noise scales; the final two rows show the generated trajectory and reference MD.
    Ribbon gaps mark O3$'$--P distances of at least 2.6~\AA{}.}
    \label{fig:denoising-high}
\end{figure}

\Cref{fig:denoising-low,fig:denoising-medium,fig:denoising-high} show stepwise trajectory generation for low-, median-, and high-motion held-out RNAs selected by reference-MD C1$'$ RMSD percentiles.

\end{document}